\documentclass{article}
\usepackage{hyperref}

\usepackage[accepted]{icml2024}

\usepackage{xspace}

\usepackage[utf8]{inputenc} 
\usepackage[T1]{fontenc}    
\usepackage{hyperref}       
\usepackage{url}            
\usepackage{booktabs}       
\usepackage{amsfonts}       
\usepackage{nicefrac}       
\usepackage{microtype}      
\usepackage{xcolor}         
\usepackage{esvect} 
\usepackage{subcaption}
\usepackage{tikz}

\usepackage{algorithm}
\usepackage{algorithmic}
\usepackage[utf8]{inputenc} 
\usepackage[T1]{fontenc}    
\usepackage{hyperref}       
\usepackage{url}            
\usepackage{booktabs}       
\usepackage{amsfonts}       
\usepackage{nicefrac}       
\usepackage{microtype}      
\usepackage{xcolor}         
\usepackage{pifont}
\usepackage{amssymb}

\usepackage{add_style}
\usepackage{bbm}
\usepackage{cleveref}

\newcommand{\penghang}[1]{{\color{purple} #1}}

\allowdisplaybreaks

\begin{document}

\twocolumn[
\icmltitle{
LLM-driven Imitation of Subrational Behavior : Illusion or Reality?
}

\icmlsetsymbol{equal}{*}
\begin{icmlauthorlist}
    \icmlauthor{Andrea Coletta}{a}
    \icmlauthor{Kshama Dwarakanath}{b,equal}
    \icmlauthor{Penghang Liu}{c,equal}
    \icmlauthor{Svitlana Vyetrenko}{b}
    \icmlauthor{Tucker Balch}{c}
\end{icmlauthorlist}
\icmlaffiliation{a}{Applied Research Team (ART) - IT Department, Bank of Italy, Rome, Italy. This research work was carried out when Andrea Coletta was employed at J.P. Morgan AI Research. The views expressed in this paper are those of the authors and do not necessarily reflect those of the Bank of Italy. }
\icmlaffiliation{b}{J.P. Morgan AI Research, California, USA.}
\icmlaffiliation{c}{J.P. Morgan AI Research, New York, USA}

\icmlcorrespondingauthor{Andrea Coletta}{andrea.coletta@bancaditalia.it}
\icmlcorrespondingauthor{Kshama Dwarakanath}{kshama.dwarakanath@jpmorgan.com}
\icmlcorrespondingauthor{Penghang Liu}{penghang.liu@jpmchase.com}

\vskip 0.3in
]
\printAffiliationsAndNotice{\icmlEqualContribution}

\begin{abstract}
Modeling subrational agents, such as humans or economic households, is inherently challenging due to the difficulty in calibrating reinforcement learning models or collecting data that involves human subjects.
Existing work highlights the ability of Large Language Models (LLMs) to address complex reasoning tasks and mimic human communication, while simulation using LLMs as agents shows emergent social behaviors, potentially improving our comprehension of human conduct.
In this paper, we propose to investigate the use of LLMs to generate synthetic human demonstrations, which are then used to learn subrational agent policies though Imitation Learning.
We make an assumption that LLMs can be used as implicit computational models of humans, and propose a framework to use synthetic demonstrations derived from LLMs to model subrational behaviors that are characteristic of humans (e.g., myopic behavior or preference for risk aversion). 
We experimentally evaluate the ability of our framework to model sub-rationality through four simple scenarios, including the well-researched ultimatum game and marshmallow experiment. 
To gain confidence in our framework, we are able to replicate well-established findings from prior human studies associated with the above scenarios. 
We conclude by discussing the potential benefits, challenges and limitations of our framework. 
\end{abstract}



\section{Introduction}
In the recent years, Large Language Models (LLMs) have been among the most impressive achievements in Artificial intelligence (AI) and Machine Learning (ML), raising questions about whether we are close to an Artificial General Intelligence (AGI) system~\cite{bubeck2023sparks,fei2022towards}. Foundation models like ChatGPT~\cite{openai_chatgpt} have shown remarkable performance across a wide range of novel and complex tasks, including coding, vision, healthcare, law, education, and psychology~\cite{bubeck2023sparks}. These models perform close to human experts, without needing additional re-training or fine-tuning, apparently mimicking human abilities and reasoning~\cite{guo2023close,min2021recent,bubeck2023sparks}. In particular, recent work has demonstrated how LLMs can emulate human reasoning in solving complex problems: \textit{LLMs can simulate chain of thoughts, generating a reasoning path to decompose complex problems into multiple easier steps to solve}~\cite{kojima2022large,wei2022chain,ho2022large}.
We could conclude that LLMs are able to replicate high-level cognitive and reasoning capabilities of humans~\cite{huang2022towards,hagendorff2022machine}, 
especially with bigger foundation models like GPT-4~\cite{wei2022emergent}.
If we assume this is true, a natural question arises: \textbf{``Can we leverage LLMs to simulate aspects of human behavior that deviate from perfect rationality so as to eventually improve our understanding of diverse human conduct?''}

Modeling human behavior has seen significant interest in various domains, ranging from robotics for human-robot collaboration~\cite{dragan2015effects}, to finance and economics for modeling human investors and consumers~\cite{liu2022biased,akerlof2010animal}. In studies of human-robot interactions, it is commonly embraced that humans must not be modeled as optimal agents~\cite{carroll2019utility}, and an irrational human when correctly modeled, can communicate more information about their objectives than a perfectly rational human can~\cite{chan2021human}. 
For instance, humans exhibit bounded rationality~\cite{simon1997models}, often making satisfactory but not optimal decisions due to their limited knowledge and processing power.
They are also known to be myopic in their decision making as they care more about short-term rewards than long-term rewards. 
While the discount factor in reinforcement learning (RL) may capture myopicity through exponential discounting, there is ample evidence to show that other temporal discounting models such as hyperbolic discounting~\cite{ainslie1992picoeconomics} and quasi-hyperbolic discounting~\cite{laibson1994hyperbolic} are more realistic in capturing time-inconsistent human behavior~\cite{chabris2010intertemporal}. 
The latter models are of particular interest in modeling household consumption and saving behavior in economic studies~\cite{angeletos2001hyperbolic}.
Empirical evidence from humans also demonstrates that the rate of discounting reduces with an increase in the amount of reward~\cite{green1997rate}. When combined with the variability of the rate parameter across humans, this makes the calibration of parameters challenging for different humans, reward amounts, and scenarios.
For example, works on estimating parameters for temporal discounting models from human studies exhibit disagreement among each other due to additional experiment-specific considerations~\cite{frederick2002time}. 
The non-exponential discounting models, along with the dependence of discount factor on the reward pose challenges to classical RL algorithms relying on the validity of the Bellman equation.
Other models have been proposed to encompass different human biases, including prospect theory, illusion of control, and optimism, which often employ nonlinear regression techniques to calibrate their parameters~\cite{langer1975illusion,tversky1992advances,sharot2007neural}.
However, such approaches demand substantial effort in collecting samples of human behavior, and it is impractical to obtain a set of parameters capable of representing individuals with different levels of bias and rationality. Also, recruiting human subjects for behavioral experiments comes with concerns surrounding ethics, privacy and general subject disinterest in the absence of monetary incentives~\cite{desposato2015ethics,abeler2015self}.

Recent work studies the possibility of using LLMs to simulate and analyze human behaviors~\cite{ziems2023can,chiang2023can,aher2023using,sejnowski2023large,horton2023large,argyle2023out,korinek2023language,williams2023epidemic}.
G. Aher et al.~\cite{aher2023using} evaluate how LLMs effectively simulate different aspects of human behavior, reproducing findings from some classic economic and psychology experiments.
J. J. Horton~\cite{horton2023large} investigates the computational capabilities of GPT-3 to model economic agents according to classical behavioral economics literature. 
Other work employs interactive LLM agents to simulate more sophisticated societal contexts, where they are eventually able to reproduce realistic social interactions and emergent behaviors~\cite{park2023generative,park2022social,gao2023s}.
Such remarkable performance in agent-based simulation could be related to the ability of LLMs to emulate different individuals, as studied in~\cite{argyle2023out}. Their work shows how LLMs effectively emulate different sociodemographic human groups under appropriate conditioning procedures.
Similarly, P. Schramowski et al.~\cite{schramowski2022large} discover that LLMs, because of how they are trained and designed, eventually contain human-like biases reflecting our behavior and morality. In fact, LLMs naturally encode a wide range of human behavior seen in the training data~\cite{bommasani2021opportunities,brown2020language}, potentially representing implicit computational models of humans.
%
The ability to display such human qualities is what makes LLMs a disruptive tool, possibly leading to a new \textit{paradigm shift}: scientists could be able to easily conduct new experiments that previously would have been time-consuming or just impossible, by using LLMs for simulation~\cite{kuhn1964structure,ziems2023can}.

In this work, we investigate the ability of LLMs to synthesize a broad range of subrational human behavior in temporal decision making tasks. Our framework learns behavioral policies through Imitation Learning (IL)~\cite{hussein2017imitation} using synthetic human-like demonstrations generated by LLMs. This addresses the limitations of classical RL techniques in modeling subrational behavior~\cite{ng2000algorithms} by overcoming the need for human demonstration data, which is expensive to gather for different human \textit{subgroups}.
Our approach assumes that LLMs represent implicit computational models of humans. Therefore, we can generate synthetic demonstrations to model subrational human behaviors, including irrational and myopic behaviors. 
Our primary goal is to initiate a novel research area that uses LLMs to synthesize subrational behavior, which has received limited preliminary attention~\cite{aher2023using,horton2023large,korinek2023language} albeit with enormous potential. Specifically, we summarize our contributions as follows:
\\
$\bullet$ We propose a novel research direction of using LLMs to calibrate subrational decision making models;
\\
$\bullet$ We propose a framework to leverage the ability of LLMs to generate synthetic demonstrations in order to learn subrational agent policies through IL;  
\\
$\bullet$ We evaluate the ability of our framework to produce subrational behavior as observed in established human studies;
\\
$\bullet$ Finally, we discuss the potential benefits, opportunities and pitfalls of utilizing LLMs for human modeling, outlining future research directions in this domain.

\section{Literature Overview}
Deep learning models have recently found extensive application in simulating complex systems and analyzing agent behaviors~\cite{ha2018world,matsuo2022deep}. They have been employed in various domains, including complex financial markets~\cite{coletta2022learning, coletta2023k}, as well as urban driving scenarios~\cite{hu2022model}. 
The advent of LLMs has created more opportunities for modelling complex systems, especially focused on human behaviors. 
Recent research efforts like ~\cite{park2023generative} and \cite{gao2023s} have utilized LLMs to create sophisticated agent-based simulators. These simulators contain LLM based agents that interact with others through natural language prompts carefully designed by incorporating world perception, reflection and planning within a defined environment. They demonstrate the ability of agents to generate \textit{believable} behavior as assessed by human evaluators, while giving rise to emergent outcomes such as information diffusion,
relationship formation, and agent coordination. 
\\
J.J. Horton~\cite{horton2023large} demonstrated the ability of LLMs to mimic human preferences arising from different qualitative personas. For example, given the following scenario \textit{``A hardware store has been selling snow shovels for \$15. The morning after a large snowstorm, the store raises the price to \$20.''}, the "libertarian" LLM found the increase acceptable while for the "socialist" LLM the increase is unfair. 
Also, recent work~\cite{argyle2023out,aher2023using} validates the ability of LLMs to model humans from a variety of subgroups. 
\\
%
%
This prompts us to consider whether LLMs can effectively capture subrational human behaviors without the need for extra effort in reward design~\cite{chan2021human}, eventually supporting or replacing traditional agent-based models~\cite{lebaron2006agent,bouchaud2018trades}.
\\
Instead of modelling an entire system using LLM agents, which is expensive and harder to control, we use LLMs to generate a few synthetic demonstrations for subrational behavior. Such demonstrations can be validated and used within an IL framework to learn agent policies encompassing existing human biases and beliefs. Complementary work in~\cite{kwon2022reward} employs LLMs to directly generate a proxy reward function based on natural language prompts. 
While such a proxy reward can align the training of policies with user-defined objectives, it cannot generate human behavior beyond what is reflected in the reward function. For example, \cite{reddy2018you} argues that subrational human behavior can arise from incorrect beliefs of environment dynamics rather than the reward function. Also, such a proxy reward does not handle non-exponential discounting models, such as hyperbolic or quasi-hyperbolic discounting. 
By directly updating the policy network to mirror human-like behavior generated through LLMs, our methodology addresses these limitations, bypassing the dependence on exponential MDPs. On the other hand, we note that our work requires more careful generation of synthetic demonstrations covering the state space of interest. A lacking of which coverage can be handled using approaches combining RL and IL~\cite{correia2023survey}. 
\\
This new paradigm of generating synthetic human demonstrations using LLMs could benefit the broader research community. LLMs can compensate for the scarcity of expert demonstrations, which are extremely useful to learn behaviors from~\cite{hester2018deep} and calibrate human subrationality models proposed in psychological and economic studies~\cite{tversky1992advances,green1997rate,simon1997models}.

\section{Learning Subrational Behavior using LLMs}
We here discuss our framework to learn subrational policies using LLM demonstrations. We first generate synthetic human demonstration data prompting the LLMs, and then use IL to train agent policies based on such demonstrations.
We use Markov Decision Processes (MDPs) as a natural underlying decision model for an agent in temporal decision making tasks. An MDP is a tuple $\mathcal{M} = (\mathcal{S}, \mathcal{A}, \mathcal{P}, \mathcal{R})$ comprising: the state space $\mathcal{S}$; the action space $\mathcal{A}$; a transition function $\mathcal{P}: \mathcal{S} \times \mathcal{A} \rightarrow \Delta(\mathcal{S})$, where $\Delta(\mathcal{S})$ denotes probability distributions over state space $\mathcal{S}$; and a reward function $\mathcal{R}: \mathcal{S} \times \mathcal{A} \rightarrow \mathbb{R}$. 
A rational agent follows policy $\pi:\mathcal{S}\rightarrow \Delta(\mathcal{A})$ that maximizes the expected reward:
\begin{equation}
J(\pi) = \mathbb{E}_{a_t\sim\pi(s_t),s_{t+1}\sim\mathcal{P}(s_t,a_t)}\left[\sum_{t=0}^{T-1} \gamma^t \mathcal{R}(s_t, a_t)  \right] \nonumber
\end{equation}
where $\gamma \in [0,1]$ represents the discount factor that is used to prioritize immediate rewards over those got later in time. 
\\
Three challenges of using direct RL to model human behavior are: 1) the design of a reward function $\mathcal{R}$ that accurately captures desired human behavior~\cite{ng2000algorithms,kwon2022reward}; 2) the large amount of data and training required before RL policies reach reasonable performance~\cite{hester2018deep}; and 3) the inability to capture time-inconsistent human preferences when using exponential discounting of rewards e.g., preference reversals~\cite{chabris2010intertemporal}.  
We propose to mitigate these challenges by using a small set of synthetic human (or expert) demonstrations that are generated using LLMs.
Such demonstration data can overcome the need to design an intricate reward function as we can use IL. 

\begin{algorithm}[t]
    \caption{\textit{IL with LLM Demonstrations} }
    \label{alg:dqn_with_expert_demonstration}
    \begin{algorithmic}
        \STATE \textbf{Input:} Demonstrations $\mathcal{D} = \{(s_i, a_i)\}_{i=1}^{n}$, 
        epochs $E$
        \STATE \textbf{Initialize:} Policy weights $\Theta$
        \FORALL{$k$ from 1 to $E$}
            \STATE \textbf{Sample} mini-batch $\boldsymbol{s} \in \mathcal{D}\ $    
                \STATE Human-preference $h(\boldsymbol{s}) = $ KDE$(\{\boldsymbol{a} | (\boldsymbol{s}, \boldsymbol{a}) \in \mathcal{D}\})$
                \STATE Policy preference $n(\boldsymbol{s})=\text{softmax}\left(Q_{\Theta}(\boldsymbol{s},\boldsymbol{a})\right)$
                \STATE Compute loss $||h(\boldsymbol{s}) - n(\boldsymbol{s})||^2$
                \STATE Perform gradient descent update on $\Theta$
        \ENDFOR
    \end{algorithmic}
\end{algorithm}

\subsection{Generation of LLM Demonstrations}  
The synthetic demonstration data comprises state action pairs over numerous episodes $\mathcal{D}=\lbrace(s_i,a_i)\rbrace_{i=1}^n$, and is generated by prompting the LLM using either chain-of-thoughts to create a reasoning path~\cite{wei2022chain}, or summarizing the state and possible actions using prompt engineering~\cite{reynolds2021prompt}. We report details of data generation in the experimental section where we use GPT-4~\cite{openai2023gpt4} due to its expressivity.
\\
Demonstration data can massively accelerate the RL training process when initialized with a policy learned using supervised learning on the data. 
Importantly, human agents may have irrational behavior in specific scenarios (i.e., states) which may be difficult to identify and model. LLMs have shown the ability to reproduce such biased behaviors~\cite{aher2023using,schramowski2022large} providing credible and novel advantages to using such demonstrations to model subrational human agents. 
Lastly, these demonstrations can be used to fine-tune exponential RL policies to capture time-inconsistent human behavior, which would have otherwise been intractable to model using standard RL.

\subsection{Imitation Learning with LLM Demonstrations}
Given demonstration data $\mathcal{D}$, we can use IL to learn a policy $\pi_{\theta}$ for the agent, formally defined as:
\begin{equation}
    \argmax_{\theta} \mathbb{E}_{(s,a)\sim\mathcal{D}} [ \log\pi_{\theta} (a | s) ]\nonumber
\end{equation}
Algorithm~\ref{alg:dqn_with_expert_demonstration} reports the proposed procedure, where we initialize the agent policy network $\pi_{\Theta}$, and in particular the value function $Q(s,a)$. Then for each state $s \in \mathcal{S}$ we extract the distribution of agent actions in the synthetic LLM demonstrations, using a non-parametric Kernel Density Estimation. We then update the policy $\pi_{\Theta}$ by minimizing the difference between its estimated action distribution and the one from synthetic demonstrations.  While more sophisticathed algorithms can be employed (e.g., GAIL~\cite{ho2016generative}) we adopt a simple algorithm as our goal is to study the application of LLMs in generating synthetic human demonstrations, rather than to propose a novel IL algorithm.
Note that pure IL like above requires demonstrations encompassing the entire state space. If not, one could start with learning the policies using an RL training phase followed by IL, similarly to~\cite{hester2018deep}.

\section{Experiments}
In this section, we test the ability of our framework to simulate human behaviors in four economic games where humans demonstrate subrational behavior, and show that we can replicate findings from existing literature.
We use GPT-4~\cite{openai2023gpt4} as the current most expressive model, and a 3-layer neural network for the $Q$ value function. We consider the following games: 
\\
$\bullet$ The \textbf{Ultimatum Game}~\cite{guth1982experimental}, a game theory experiment where a player decides how to split an amount of money, and another player can accept or reject the split. 
 \\
$\bullet$ The \textbf{Stanford marshmallow experiment}~\cite{mischel1970attention} where a player is offered a choice between one small but immediate reward, or a bigger reward if they waited for a period of time.
\\
$\bullet$ The \textbf{Double or nothing gamble} where a player is offered a choice to double their money or lose everything, in order to illustrate human decisions under risk as described by prospect theory~\cite{kahneman1979prospect}.
\\
$\bullet$ The \textbf{Procrastination experiment}~\cite{o1999doing} where a student player chooses which day to write a report and miss a good movie given an academic deadline. 

\subsection{Ultimatum Game}\label{sec:ultimatum}

\paragraph{Scenario.} The ultimatum game considers two players splitting a given amount of money $T$. The first player (the proposer) decides to split the money between himself and the responder. The responder (who knows the total sum) decides to either accept or reject the split. If the responder rejects the proposal, both players receive nothing; if the responder accepts, they both get the agreed amount of money. For example, the proposer decides to split $T=\$10$ by giving $\$2$ to the responder and taking $\$8$ for himself; if the responder accepts they get $\$8$ and $\$2$, respectively. Otherwise they both get nothing. 

\paragraph{Social Preferences and Limited Rationality.}
If we assume that agents have stable, well-defined, and rational preferences, we realize that the
theory for the simple ultimatum game is rather obvious: a rational responder would accept any non-zero proposal. In fact, even if the proposal is unfair, getting something is better than getting nothing. This represents a Nash equilibrium for the ultimatum game. 
In reality though, human behavior is much different. The work of Nobel laureate Richard Thaler~\cite{thaler1988anomalies} focuses on some "anomalies" in the ultimatum game, defined as demonstration of limited rationality and social preferences by human participants. They show that a human responder will likely reject unfair proposals, as they prefer to "punish" the proposer, even though they will receive nothing. According to their own social preference, people will prefer "fair" (i.e., 50:50) splits~\cite{sanfey2003neural}, or in general they often reject offers less than 30\% of the initial sum $T$~\cite{HOUSER201419,KRAWCZYK2018283}.

\begin{figure}[t]
\centering
\subcaptionbox{\scriptsize RL}{\includegraphics[width=0.44\linewidth]{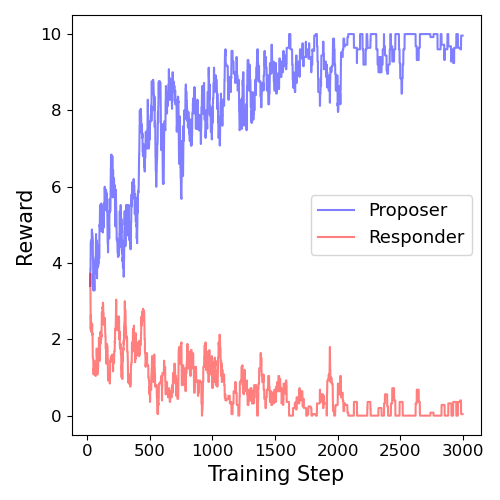}}\hspace{0.05\linewidth}
\subcaptionbox{\scriptsize LLMs (human)}{\includegraphics[width=0.44\linewidth]{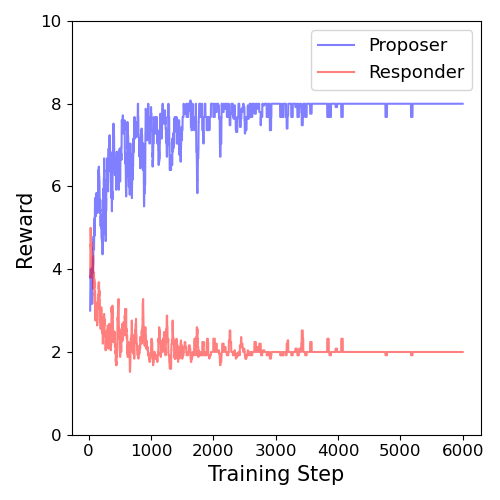}}
\\
\subcaptionbox{\scriptsize LLMs (fair)}{\includegraphics[width=0.44\linewidth]{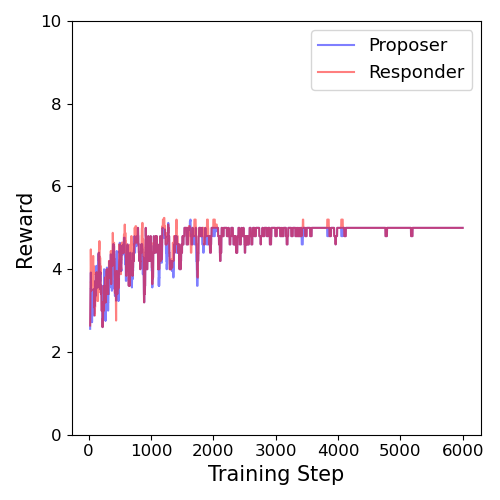}}\hspace{0.05\linewidth}
\subcaptionbox{\scriptsize Proposal Distribution}{\includegraphics[width=0.44\linewidth]{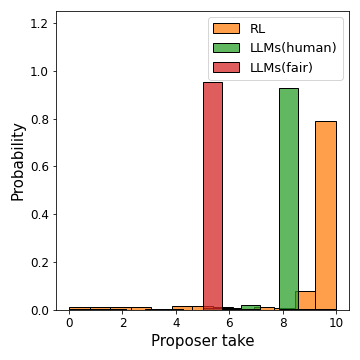}}
\caption{Ultimatum Game experiment. 
}\label{fig:ultimatum_fair}
\end{figure}

\paragraph{Experimental Setting.}
We consider an initial amount of $T=\$10$ for the game. Therefore, the game has 10 possible actions for the proposer $x \in \{0, \ldots, 10\}$, where $x$ is the amount to offer and ($10 - x$) is the amount they keep for themselves. Accordingly, each offer is a possible state for the responder who has 10 states and 2 actions $a \in $\{\textit{accept}, \textit{reject}\}. In this experimental setup, we focus on modelling the human responder, which we consider the more interesting player in the game. We model the proposer using a Multi-Armed Bandit~\cite{sutton2018reinforcement}; and use an RL policy network for the responder. The two agents are trained together using as reward the amount of money they individually get. We alternately use IL with LLM demonstrations for a \textbf{human responder} or one that cares about \textbf{a fair split} to obtain the responder policy.

\paragraph{Synthetic demonstrations collected using LLMs.} 
We instruct GPT-4 as \textit{"Impersonate Jerry who cares a lot about fairness."} or \textit{"Impersonate a human called Jerry"}. We call the two experiments \textit{"Fair"} and \textit{"Human"}, respectively. We then prompt the model as: {\em John receives \$T, and he proposes to offer Jerry x and keep T-x  for himself. If Jerry accepts they both get the agreed amount, but if Jerry rejects they both receive nothing. Continue the sentence: `Jerry decides to'}, where $T=10$ and $x \in \{0, \ldots, T\}$. We then convert each prompt and answer into a state-action $(s, a)$ pair. We generate 10 synthetic observations for each state $x$.

\paragraph{Results.}
Figure~\ref{fig:ultimatum_fair} shows rewards and distribution of money kept by the proposer. 
Figure~\ref{fig:ultimatum_fair}.(a) shows classic RL training for both agents, where the proposer learns to split the money in an unfair manner, and the responder accepts any splits. In fact, this is the expected outcome from a fully rational responder. 
In Figure~\ref{fig:ultimatum_fair}.(b) we use LLM demonstrations to train a "human" responder model, which now rejects very low (i.e., less than 20\%) splits, and thus the two agents converge to a \$8 and \$2 split. In Figure~\ref{fig:ultimatum_fair}.(c) we use LLM demonstrations to train a "fair" responder model, which now rejects unfair (i.e., unbalanced) splits, and thus the two agents converge around a 50:50 split. Finally, Figure~\ref{fig:ultimatum_fair}.(d) reports the distributions of money the proposer keeps for the three variants. 

\paragraph{Comparison to human subject studies.}
We compare our results with those in economic literature
in~\cite{HOUSER201419} and~\cite{KRAWCZYK2018283}. These works show that a human responder is less likely to accept offers below \$3, out of \$10. This is inline with our results confirming that synthetic LLM demonstrations could be effectively used to model human interactions, as is also highlighted in recent work~\cite{aher2023using}. Importantly, LLMs enable the simulation of a broader range of scenarios without any additional costs (e.g., \$100 instead of \$10).

\subsection{Stanford Marshmallow Experiment.}\label{sec:stanford_exp}

\paragraph{Scenario.} The Stanford Marshmallow Experiment is a classic psychological study that explores delayed gratification in children. In this experiment, a child is placed in a room with a tempting marshmallow or a similar treat. The researcher presents the child with a choice: they can either eat the marshmallow immediately or wait for a predetermined amount of time (usually 15 minutes) without eating it. If the child can resist the temptation and wait until the time elapses, they are rewarded with an additional marshmallow. This experiment is used to investigate the ability of children to delay their desires and make choices based on long-term benefits rather than immediate gratification.

\paragraph{Self-Control and the Stanford Marshmallow Experiment.}
If we assume that children have strong self-control and can easily delay gratification, the theory behind the Stanford Marshmallow Experiment might seem straightforward: a child with impeccable self-discipline would wait patiently for the delayed reward, in this case, an extra marshmallow. In this ideal scenario, the child would understand that waiting for the second marshmallow is a more desirable outcome than consuming the first one immediately, representing a form of self-control equilibrium in this experiment. In reality, however, the behavior of children in the marshmallow experiment is quite different. The work of Walter Mischel~\cite{mischel1970attention} uncovered what could be considered "anomalies" in human self-control and decision-making. They demonstrated that many children struggle with self-control and are unable to resist the immediate temptation of eating the first marshmallow, even though waiting would lead to a greater reward. In particular, such work revealed that children's self-control abilities vary, and their choices in the experiment are influenced by factors like age, environment, and individual differences. While incorporating these factors into RL is non-trivial, obtaining human demonstrations for very specific settings can be expensive or impossible, and synthetic demonstrations from LLMs could overcome these challenges.

\paragraph{Experimental Setting.}
We focus on modelling a child aligned with the psychological game in this setup. 
We consider a single agent with 2 actions $a \in \{accept, wait\}$ for the candy in the current state.  We consider one initial state where the agent can choose to accept the candy now or wait, and two subsequent states derived by each choice. 

\begin{figure}[t]
\centering
\subcaptionbox{\scriptsize RL}{\includegraphics[width=0.44\linewidth]{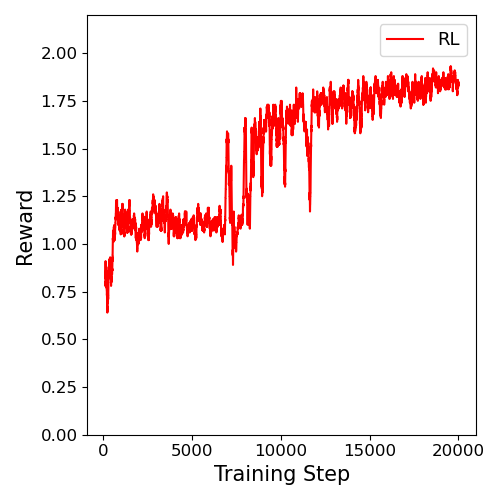}}\hspace{0.05\linewidth}
\subcaptionbox{\scriptsize Myopic RL}{\includegraphics[width=0.44\linewidth]{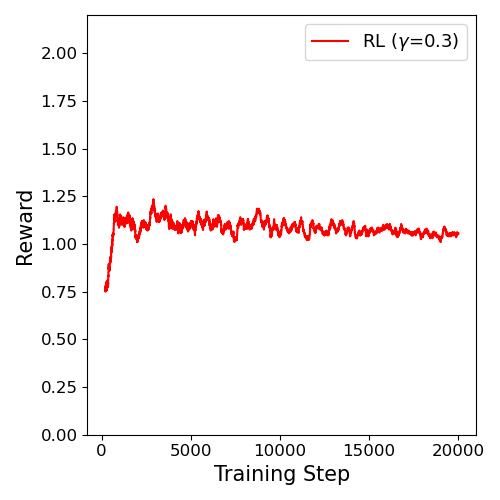}}\\
\subcaptionbox{\scriptsize LLMs}{\includegraphics[width=0.44\linewidth]{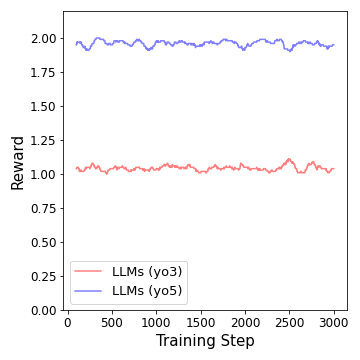}}\hspace{0.05\linewidth}
\subcaptionbox{\scriptsize Prob. of waiting}{\includegraphics[width=0.44\linewidth]{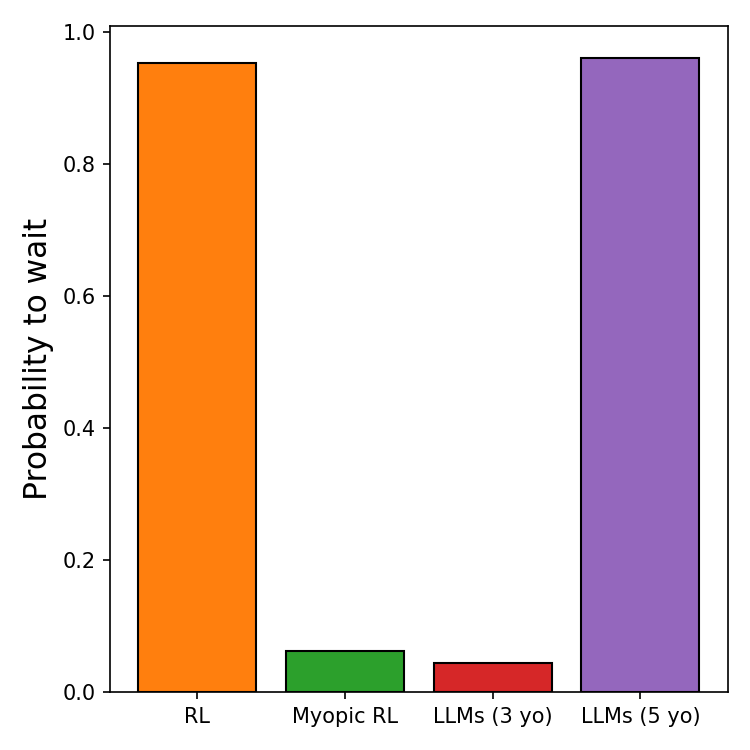}}
\caption{Stanford marshmallow experiment
}\label{fig:stanford_res}
\end{figure}

\paragraph{Synthetic demonstrations collected using LLMs.} 
We instruct GPT-4 as \textit{"You are Janice a Y years old child."}. We prompt the model as: {\em Janice is offered to get one candy now, or to wait for 2 more hours to eventually get two candies. Continue the following sentence 'Janice decides to'}, where $Y \in \{2, 5\}$. 
Interestingly, we noticed that with `15 minutes' waiting like the original experiment, the model is more inclined to wait (Appendix~\ref{tab:stanford_answers_15min}), possibly indicating different human preferences w.r.t. the original paper dated more than 50 years ago. We generate 10 synthetic observations for each age $Y$.

\paragraph{Results.}
In Figure~\ref{fig:stanford_res}.(a) we show a classic RL approach trained until convergence, where the reward is the amount of candies the agent gets. The agent is completely rational, and it learns to always wait to get two candies. Most important, it is very hard to model different background factors for the agent, like age. For example, can we derive an RL policy for a child of 2 years vs a child of 5 years? A common approach could be myopic RL modelling shown in Figure~\ref{fig:stanford_res}.(b) where we modify the discount factor $\gamma$ to let the agent being more myopic (i.e,. looking only for immediate rewards). In the setting with $\gamma=0.3$ the agent chooses to take the candy immediately, with a lower reward. Even if we can use myopicity modeling, it is not clear how to relate the value $\gamma$ with the age of the child. Instead, an LLM could be easily prompted to get precise synthetic human observations. Figure~\ref{fig:stanford_res}.(c) shows that if we directly train an IL policy using LLM demonstrations, we are able to reproduce behavior of a 2-year old child who takes one candy immediately, or a more rational 5-year old child that waits for more candies. 

\paragraph{Comparison to human subject studies.} 
We compare our results with the study of waiting behavior of nursery school children by W. Mischel~\cite{mischel1970attention} comprising 16 boys and 16 girls, with age ranging from 3 to 8 years. In this experiment, they found that most children do not wait, when both rewards are visible to them; while around 75\% of children wait if both rewards having been removed from their sight. Our results show a possible reconstruction of this experiment: a 2-year old child does not wait, while a 5-year old child is more rational and waits; with a mix of both actions for a 3-year old child (details in the appendix).
These findings are aligned with the aforementioned study which suggests that the capacity to wait for long-term goals is develop markedly at about ages 3-4.
Nonetheless, we also note that the original experiments are from 50 years ago, making current human biases and needs very different (e.g., children may have easier access to treats now). 

\subsection{Double or Nothing Gamble}\label{sec:prospect_exp}

\paragraph{Scenario.}
We create a simple economic game named ``\textit{double or nothing}" to illustrate human bias in decision-making under risk. 
Assume that two people are gambling, and person A has lost an initial bet of \$5 against person B. Before the initial bet is paid, person A (or B) can choose to play a second bet of \$5 with a higher probability ($>50$\%) of losing (or winning). Since the second bet is an unfair game that favors the winner (person B), a rational agent will always deny the second bet as person A or accept the second bet as person B. However, humans in real life are willing to take the risk of the second bet as person A (hoping to offset their loss from the previous bet), or deny the second bet as person B (afraid of losing their current reward).

\paragraph{Prospect theory.}
The expected utility theory is a fundamental model of human decision making under uncertainty, which says that an agent chooses between risky prospects based on expected utility: $U = \sum^n_{i=1} p_i \cdot x_i$, where $p_i$ is the probability of an outcome with payoff $x_i$. RL aligns with this hypothesis as the value of a state is determined by the expected utility of next states weighted by their transition probabilities. But, a substantial body of evidence has shown that human decisions systematically violate this theory. To model such deviation, Kahneman and Tversky introduced prospect theory \cite{kahneman1979prospect} which has been widely used for explaining experimental findings on human risk taking over several decades \cite{benartzi1995myopic,thaler1997effect,barberis2013thirty}.
The model consists of two key elements:
(1) a biased value function $v(x_i)$ that is concave for gains, convex for losses, and steeper for losses than gains, and
(2) a nonlinear probability weighting function $w(p_i)$ which overweights small probability and underweights large probability.
Humans with prospect bias are \textit{risk-averse} over gains and \textit{risk-seeking} over losses.

\paragraph{Experimental Setting.}
We create the "\textit{double or nothing}" gamble environment with unfairness factor of the second bet $\epsilon \in \{0, 0.1, 0.2, 0.3, 0.4\}$. The winner of the first bet has probability $\{0.5 + \epsilon\}$ to win the second bet against the loser. We train both players separately. Given $\epsilon$ as the state, the agent takes an action $a \in \{accept, reject\}$ to decide whether or not to add the second bet. A rational RL agent learns to maximize the expected reward from the two bets $U = \sum^n_{i=1} p_i \cdot x_i$, whereas the prospect biased RL agent has reward function $U = \sum^n_{i=1} w(p_i) \cdot v(x_i)$ per Eq.~\ref{eq:prospect_utility} and Eq.~\ref{eq:prospect_prob} with parameters estimated from real human data.

\paragraph{Synthetic demonstrations collected using LLMs.} 
We prompt GPT-4 as follows: 
\\
$\bullet$ \textbf{1)} \textit{You won an initial \$5 bet against Tom. Before you collect your reward, you can choose to add a second bet of \$5 with $\{0.5 + \epsilon\}$ probability to win. If you win the second bet you will double your gain from the initial bet, but if you lose you will gain nothing. Continue the following sentence `you decide to'}; \\
\\
$\bullet$ \textbf{2)} \textit{You lost an initial \$5 bet against Tom. Before you pay your debt, you can choose to add a second bet of \$5 with $\{0.5 - \epsilon\}$ probability to win. If you win the second bet you will recover your loss from the initial bet, but if you lose you will double your loss. Continue the following sentence `you decide to'}.
\\
We collect 10 demonstrations for each $\epsilon \in \{0, 0.1, 0.2, 0.3, 0.4\}$ and show the probability of accepting the second bet as the winner/loser in Table~\ref{tab:tom_jerry}.

\begin{table}[]
    \centering
    \caption{Prob. of a second bet in LLM demonstrations.}
    \begin{tabular}{cccccc}
         \hline
         $\epsilon$ & 0 & 0.1 & 0.2 & 0.3 & 0.4 \\ \hline
         Winner & 0.3 & 0.5 & 1 & 1 & 1\\
         Loser & 1 & 1 & 0.6 & 0 & 0\\\hline
    \end{tabular}
    \label{tab:tom_jerry}
\end{table}

\begin{figure}[t]
\centering
\subcaptionbox{\scriptsize Winner Rewards}{\includegraphics[width=0.32\linewidth]{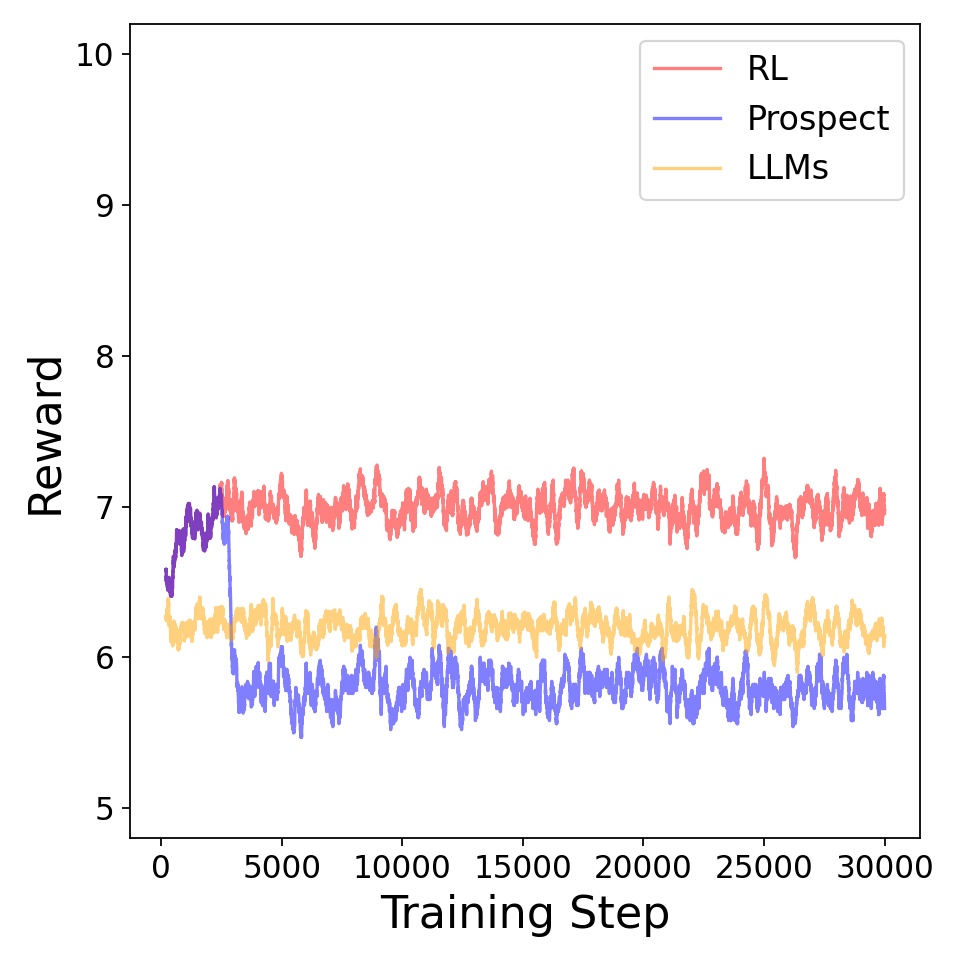}}
\hfill
\subcaptionbox{\scriptsize Loser Rewards}{\includegraphics[width=0.32\linewidth]{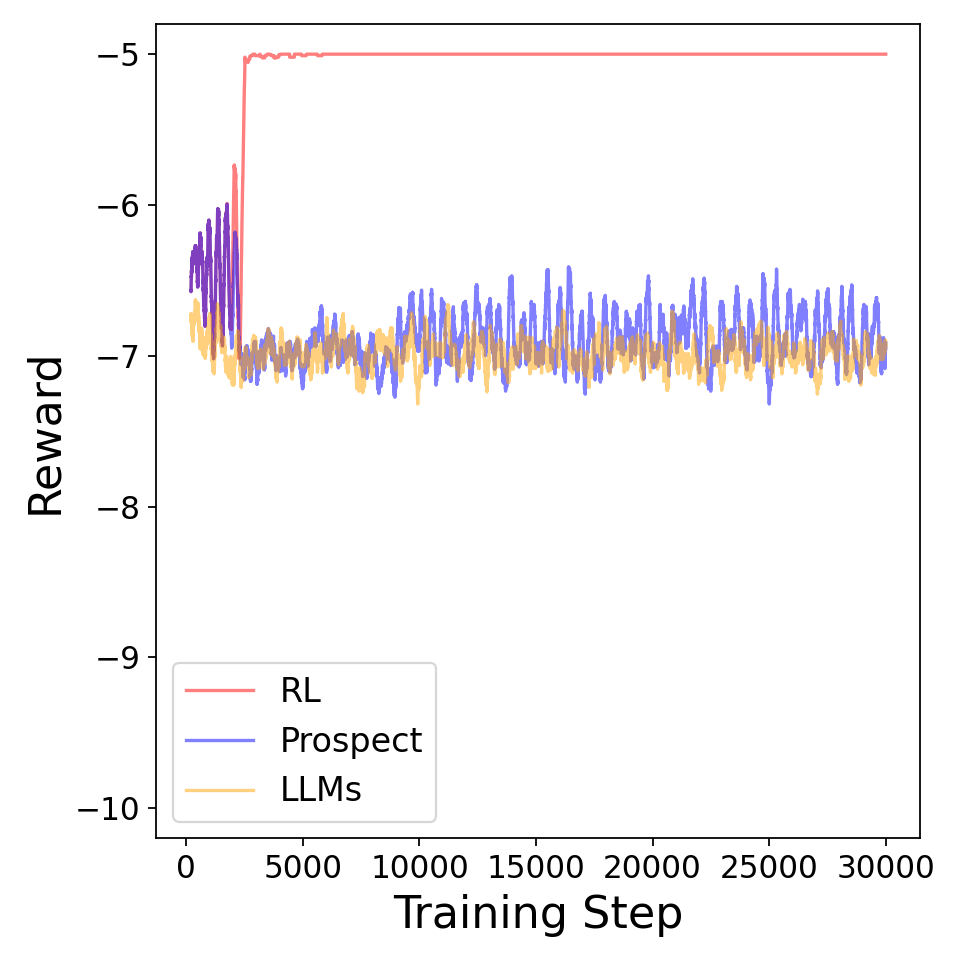}}
\hfill
\subcaptionbox{\scriptsize Prob. to accept}{\includegraphics[width=0.33\linewidth]{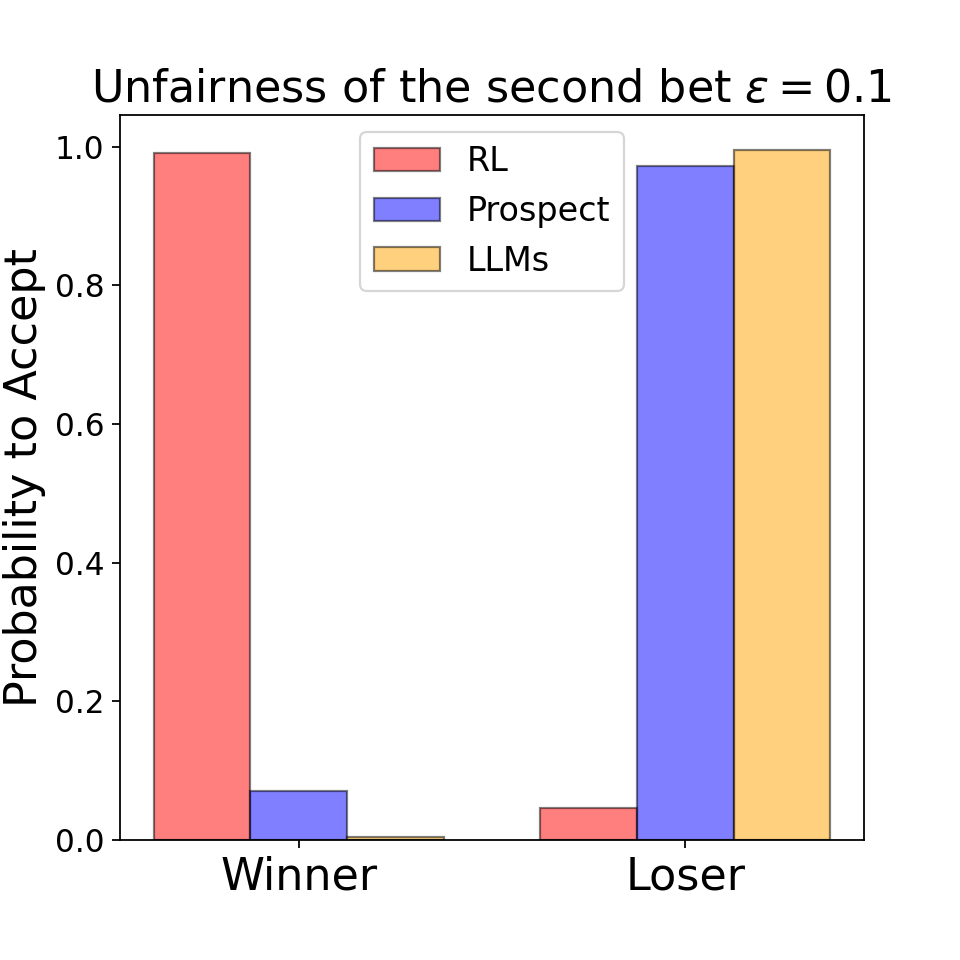}}
\caption{
"\textit{Double or Nothing}" gamble.
}
\label{fig:2bets_res}
\end{figure}

\paragraph{Results.}
Table~\ref{tab:tom_jerry} shows the probability of adding the second bet in LLM demonstrations. 
In general, we observe that the loser (winner) of the first bet is more (less) willing to take the risk of playing an additional bet, unless the unfairness of the second bet is large enough ($\epsilon \geq 0.3$).
We compare an IL policy trained over LLM demonstrations with a rational RL policy and the prospect-biased RL policy in \Cref{fig:2bets_res}. Since the second bet always favors the winner, the rational RL agent learns to maximize their reward by accepting it as winner and rejecting as loser. On the other hand, agents trained using prospect theory and LLM demonstrations yield similar (lower) rewards which are subrational e.g., when $\epsilon = 0.1$, both are not willing to take the risk as winner, but prefer to take the additional bet as loser. 

\paragraph{Comparison to human subject studies.} 
We compare behavior generated by the IL policy using LLM demonstrations to the RL policy using the prospect theory model specified by parameters estimated from real humans in the literature \cite{tversky1992advances}. Both models yield similar behavior and rewards, showing risk-averse behavior in gains and risk-seeking behavior in losses. 
These results highlight the simplicity of our framework in incorporating subrational and biased behavior into agent policies without the need to design intricate or ad-hoc reward functions.

\subsection{Academic Procrastination with Deadlines}\label{subsec:procrastination}
\paragraph{Scenario.}
A student needs to spend one day to write a course report due within the next $H$ days. They receive a credit for the submitted report on the last day. The theatre is playing movies in increasing order of quality over the next $H$ days. Which day does the student choose to write the report and miss the movie? Let $c_t$ represent the cost of missing a movie on day $t\in\lbrace1,2,\cdots,H\rbrace$ so that $0<c_1<c_2\cdots<c_H$. And, $R>0$ is the reward received upon submitting the report on day $H$. This setup is based on the example on salient costs in \cite{o1999doing}.

\paragraph{Procrastination and Quasi-Hyperbolic Discounting.}
Studies have shown that a large fraction of students procrastinate on their academic deliverables until the deadline gets very close~\cite{rothblum1986affective,day2000patterns}. This is an example for the lack of self-control observed in humans: they choose to watch a movie today with the expectation that their future self would write the report tomorrow. But when tomorrow arrives, the same decision repeats until the deadline gets very close. Sophisticated humans who are aware about their lack of self-control expect their future selves to procrastinate and complete the report earlier than naive humans who are optimistic about their future selves \cite{pollak1968consistent}. Such behavior is modeled by quasi-hyperbolic discounting \cite{laibson1994hyperbolic} where the value function in a state $s_t$ is given by 
$$V(s_t)=\mathbb{E}\left[\mathcal{R}(s_t)+\beta\sum_{k=1}^{H-t}\delta^{k}\mathcal{R}(s_{t+k})\right]$$ where $\delta\in[0,1]$ is the long-term discount factor similar to exponential discount factor $\gamma$, and $\beta\in[0,1]$ captures `short-term impatience' representing the preference over immediate rewards today vs tomorrow \cite{o1999doing}.
 With the assumption of exponential discounting (as in standard RL), one arrives at time-consistent preferences.
Time-inconsistent preferences are those where the optimal policy from today's perspective conflicts with the optimal policy from tomorrow's perspective \cite{laibson1994hyperbolic}.
While it is unclear how to modify standard RL algorithms (that rely on the Bellman equation) to arrive at this time-inconsistent human behavior modeled by quasi-hyperbolic discounting, we show next that such behavior can be generated through IL using synthetic human demonstrations from LLMs.

\paragraph{Experimental Setting.}
Let the deadline be $H=4$ with the theatre playing a mediocre movie on day 1, a good movie on day 2, a great movie on day 3 and an excellent movie on day 4. We collect LLM demonstrations for three types of students as categorized by their Grade Point Average (GPA). As a baseline for comparison, we use a policy learned using RL with costs $c_1=1$, $c_2=2$, $c_3=4$, $c_4=7$ and final reward $R=14$ with exponential discount factor $\gamma=1$.

\begin{figure}[t]
\centering
\subcaptionbox{\scriptsize LLM Demonstrations}{\includegraphics[width=0.42\linewidth]{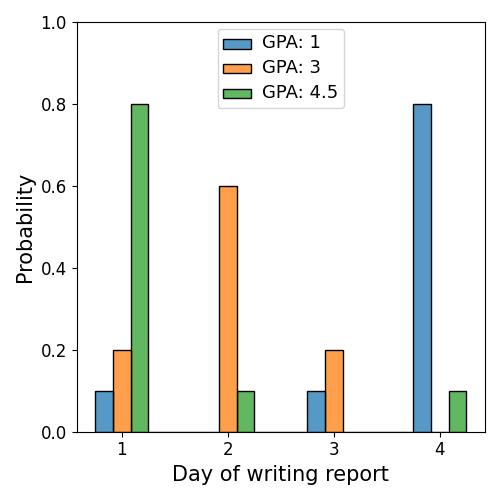}}\hspace{0.05\linewidth}
\subcaptionbox{\scriptsize RL}{\includegraphics[width=0.42\linewidth]{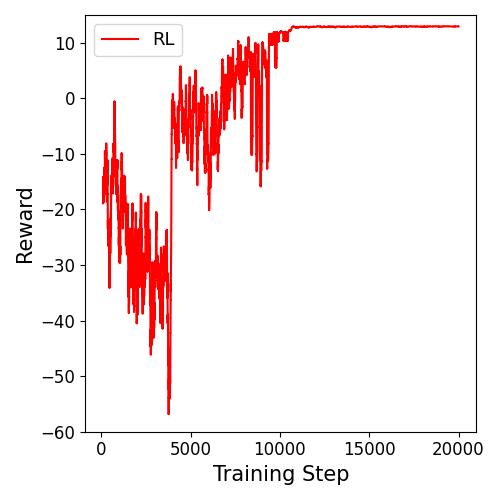}}\\
\subcaptionbox{\scriptsize LLMs}{
\includegraphics[width=0.42\linewidth]{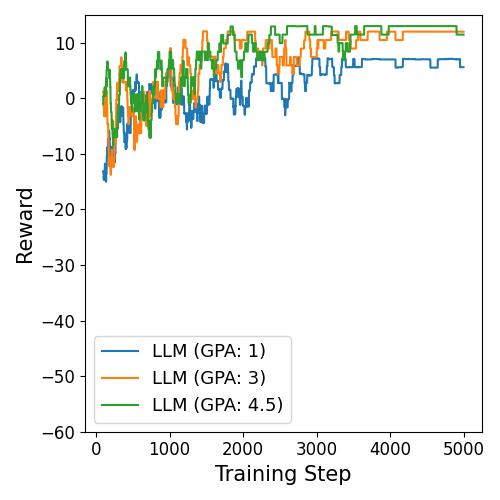}
}\hspace{0.05\linewidth}
\subcaptionbox{\scriptsize Day of writing report}{
\includegraphics[width=0.42\linewidth]{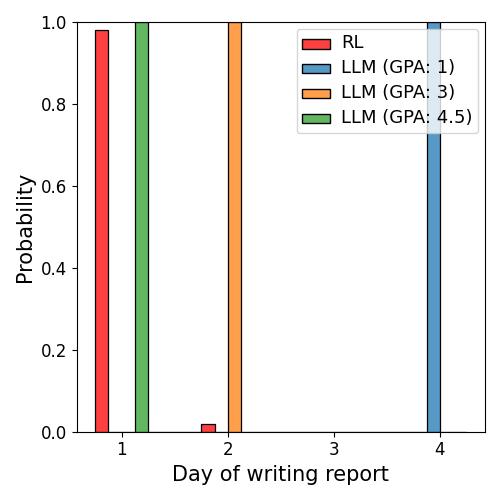}
}
\caption{
Procrastination experiment. 
}\label{fig:procrastination_4}
\end{figure}

\paragraph{Synthetic demonstrations collected using LLMs.} 
We instruct GPT-4 as \textit{"You impersonate a student with GPA $x$, who loves watching movies. GPA measures the students commitment towards their academics."}. We then prompt the model as: {\em "The theatre has a line up of increasingly better movies as days pass. The student has a deadline to submit a course report within the next four days. The student needs to pick one day to write the report, which means they will miss the movie on that day. Continue the sentence `The student writes the report on day '"}, where $x\in\lbrace1,3,4.5\rbrace$. We collect 10 demonstrations for each GPA $x$ as shown in Figure~\ref{fig:procrastination_4}.(a), where we see that the probability of writing the report on the first day increases as the GPA increases.

\paragraph{Results.} Figure~\ref{fig:procrastination_4}.(b) shows the training rewards for an RL policy using the above cost and reward setting. A histogram of the day of writing the report when using the trained RL policy is plotted in Figure~\ref{fig:procrastination_4}.(d). Observe that exponential discounting with $\gamma=1$ says it's optimal to write the report on day 1 since it picks $t$ that maximizes $R-c_t$. Thus, the RL policy always corresponds to writing the report on day 1. 
On the other hand, the LLM demonstrations can be used to train a policy by IL, whose rewards are plotted in Figure~\ref{fig:procrastination_4}.(c) with the day of writing the report in Figure~\ref{fig:procrastination_4}.(d). Note that when LLM demonstrations are used, they give rise to one policy per student type/GPA.
Observe from Figure~\ref{fig:procrastination_4}.(c) that the rewards at the end of training are lower for lower GPA values, and improve as GPA increases. 
Since the cost of writing the report increases with the day of writing it, this says that the learned IL policies give earlier days of writing the report as GPA increases. The same can be observed in Figure~\ref{fig:procrastination_4}.(d) where while the RL policy without any LLM demonstrations corresponds to writing the report on the first day, the IL policies with LLM demonstrations correspond to writing it on later days for lower GPAs. Thus, while the RL policy gives time consistent actions, incorporating LLM demonstrations helps us to capture procrastination, while accounting for differences in student attributes such as GPA.


\paragraph{Comparison to human subject studies.} 
We compare our results to a study of procrastinating behavior among 379 undergraduate students in an introductory psychology course~\cite{rothblum1986affective}. The study examines the extent of prevalence of procrastinating behavior, and the relationship between academic procrastination and academic performance. The authors found a positive correlation between self-reported procrastination and delay in taking self-paced quizzes, and a negative correlation between procrastination and grade point average. Our findings firstly capture procrastination with the help of LLM demonstrations, and secondly are in line with this study since we observe higher delays in writing the report for students with lower GPAs. However, we note that this relationship between GPA and academic delay requires deeper investigation as remarked in~\cite{solomon1984academic} since the observations are based on correlation and not causation.


\section{Discussion and Broader Context}

In this work, we investigate the utility of foundation models to synthesize subrational human decision making behavior, that deviates from perfectly rationality~\cite{simon1955behavioral}. 
Our work is motivated by the need to capture subrational and biased human behavior across disciplines ranging from economics of households~\cite{frederiks2015household} to financial markets~\cite{liu2022biased}, 
alongside the difficulty in obtaining historical data for the same to design calibrated subrational human models~\cite{tversky1992advances,green1997rate}.
We propose a framework to generate synthetic human demonstration data for a variety of human \textit{subgroups} using LLMs, that is subsequently used with IL to derive behavioral policies for each human. 
Such demonstration data has potential to be used to calibrate subrational human models while being cost-effective and efficient to generate as opposed to real human studies that cover small subsets of the human population. 
This data is also useful when trying to capture human behavior that is hard to describe through the use of a reward function~\cite{kwon2022reward}, and in cases where exponential discounting does not hold~\cite{chabris2010intertemporal}. 
Calibrated human models can improve realism of agent-based models of real economic systems, that can be subsequently used for policy applications~\cite{farmer2009economy,dong2023analyzing}.
We ground our framework by comparing our experimental findings in four economic games with real human studies in the literature, where we are able to reproduce similar findings.

As we advocate for further research in this field utilizing foundation models to capture subrational behavior in a wider class of problems beyond the games that we modeled in this work, we now discuss challenges and potential limitations to the use of LLMs for the same.
As for most applications using LLMs, we share some of the existing limitations such as the requirement for very careful prompt engineering (e.g. we end our prompts saying {\em `continue the sentence'} so as to limit responses by relevance), which may even introduce unintentional user biases into the system. Prompt engineering is particularly important when dealing with complex, temporal scenarios that are common in economics when translating the agent states into natural language. This is also accompanied by their limited ability in handling numbers that can reduce their efficiency in such scenarios~\cite{hendrycks2021measuring}. 
Also, ever since they rose to popularity, LLMs have seen constant updates accompanied by guardrails that may prevent the reproducibility of their responses. 
\\
Based on the assumption that LLMs are computational models of humans, memorization of training data by LLMs is not a limitation, as it enables us to produce human-like behavior with more fidelity. 
That said, LLM responses could be sensitive to protected attributes of agents like age, gender, ethnicity making it challenging to differentiate the effects of biases in training data vs hallucinations. 
It is also unclear if one can trust the demonstrations generated by LLMs in unseen scenarios (e.g. different ages than those seen in the marshmallow experiment). 
We perform ablation studies by modifying the prompts for the Stanford marshmallow experiment to test for reasoning (or memorization) in the appendix. 
Further, evaluation of LLM responses for their {\em human-ness} or quality can be subjective, requiring human expert knowledge. In our work, we are able to reproduce similar behavior as was observed in real human studies, some of which were conducted over 50 years ago~\cite{mischel1970attention,rothblum1986affective}. 
It is especially intriguing to think of the evolution of LLMs in the coming years as they are subsequently trained and fine-tuned on more and more LLM generated data! 

To conclude, our work shows an initial exploration of the use of LLMs for modeling subrational behavior demonstrating great potential with possible broader applications. We believe that LLMs have the capability to emerge as indispensable and cost-effective tools for modeling subrational behaviors, and help decision makers in different scenarios. This as long as their use is accompanied by a careful consideration of the challenges, limitations and potential biases. 

\section*{Acknowledgements}
This paper was prepared for informational purposes in part by the CDAO group of JPMorgan Chase \& Co and its affiliates (``J.P. Morgan'') and is not a product of the Research Department of J.P. Morgan. J.P. Morgan makes no representation and warranty whatsoever and disclaims all liability, for the completeness, accuracy or reliability of the information contained herein.  This document is not intended as investment research or investment advice, or a recommendation, offer or solicitation for the purchase or sale of any security, financial instrument, financial product or service, or to be used in any way for evaluating the merits of participating in any transaction, and shall not constitute a solicitation under any jurisdiction or to any person, if such solicitation under such jurisdiction or to such person would be unlawful. 

\bibliography{mybib}

\begin{thebibliography}{82}
\providecommand{\natexlab}[1]{#1}
\providecommand{\url}[1]{\texttt{#1}}
\expandafter\ifx\csname urlstyle\endcsname\relax
  \providecommand{\doi}[1]{doi: #1}\else
  \providecommand{\doi}{doi: \begingroup \urlstyle{rm}\Url}\fi

\bibitem[Abeler \& Nosenzo(2015)Abeler and Nosenzo]{abeler2015self}
Abeler, J. and Nosenzo, D.
\newblock Self-selection into laboratory experiments: pro-social motives versus
  monetary incentives.
\newblock \emph{Experimental Economics}, 18:\penalty0 195--214, 2015.

\bibitem[Aher et~al.(2023)Aher, Arriaga, and Kalai]{aher2023using}
Aher, G.~V., Arriaga, R.~I., and Kalai, A.~T.
\newblock Using large language models to simulate multiple humans and replicate
  human subject studies.
\newblock In \emph{International Conference on Machine Learning}, pp.\
  337--371. PMLR, 2023.

\bibitem[Ainslie(1992)]{ainslie1992picoeconomics}
Ainslie, G.
\newblock \emph{Picoeconomics: The strategic interaction of successive
  motivational states within the person}.
\newblock Cambridge University Press, 1992.

\bibitem[Akerlof \& Shiller(2010)Akerlof and Shiller]{akerlof2010animal}
Akerlof, G.~A. and Shiller, R.~J.
\newblock \emph{Animal spirits: How human psychology drives the economy, and
  why it matters for global capitalism}.
\newblock Princeton university press, 2010.

\bibitem[Angeletos et~al.(2001)Angeletos, Laibson, Repetto, Tobacman, and
  Weinberg]{angeletos2001hyperbolic}
Angeletos, G.-M., Laibson, D., Repetto, A., Tobacman, J., and Weinberg, S.
\newblock The hyperbolic consumption model: Calibration, simulation, and
  empirical evaluation.
\newblock \emph{Journal of Economic perspectives}, 15\penalty0 (3):\penalty0
  47--68, 2001.

\bibitem[Argyle et~al.(2023)Argyle, Busby, Fulda, Gubler, Rytting, and
  Wingate]{argyle2023out}
Argyle, L.~P., Busby, E.~C., Fulda, N., Gubler, J.~R., Rytting, C., and
  Wingate, D.
\newblock Out of one, many: Using language models to simulate human samples.
\newblock \emph{Political Analysis}, 31\penalty0 (3):\penalty0 337--351, 2023.

\bibitem[Barberis(2013)]{barberis2013thirty}
Barberis, N.~C.
\newblock Thirty years of prospect theory in economics: A review and
  assessment.
\newblock \emph{Journal of economic perspectives}, 27\penalty0 (1):\penalty0
  173--196, 2013.

\bibitem[Benartzi \& Thaler(1995)Benartzi and Thaler]{benartzi1995myopic}
Benartzi, S. and Thaler, R.~H.
\newblock Myopic loss aversion and the equity premium puzzle.
\newblock \emph{The quarterly journal of Economics}, 110\penalty0 (1):\penalty0
  73--92, 1995.

\bibitem[Benhabib et~al.(2010)Benhabib, Bisin, and
  Schotter]{benhabib2010present}
Benhabib, J., Bisin, A., and Schotter, A.
\newblock Present-bias, quasi-hyperbolic discounting, and fixed costs.
\newblock \emph{Games and economic behavior}, 69\penalty0 (2):\penalty0
  205--223, 2010.

\bibitem[Bommasani et~al.(2021)Bommasani, Hudson, Adeli, Altman, Arora, von
  Arx, Bernstein, Bohg, Bosselut, Brunskill,
  et~al.]{bommasani2021opportunities}
Bommasani, R., Hudson, D.~A., Adeli, E., Altman, R., Arora, S., von Arx, S.,
  Bernstein, M.~S., Bohg, J., Bosselut, A., Brunskill, E., et~al.
\newblock On the opportunities and risks of foundation models.
\newblock \emph{arXiv preprint arXiv:2108.07258}, 2021.

\bibitem[Bouchaud et~al.(2018)Bouchaud, Bonart, Donier, and
  Gould]{bouchaud2018trades}
Bouchaud, J.-P., Bonart, J., Donier, J., and Gould, M.
\newblock \emph{Trades, quotes and prices: financial markets under the
  microscope}.
\newblock Cambridge University Press, 2018.

\bibitem[Brown et~al.(2020)Brown, Mann, Ryder, Subbiah, Kaplan, Dhariwal,
  Neelakantan, Shyam, Sastry, Askell, et~al.]{brown2020language}
Brown, T., Mann, B., Ryder, N., Subbiah, M., Kaplan, J.~D., Dhariwal, P.,
  Neelakantan, A., Shyam, P., Sastry, G., Askell, A., et~al.
\newblock Language models are few-shot learners.
\newblock \emph{Advances in neural information processing systems},
  33:\penalty0 1877--1901, 2020.

\bibitem[Bubeck et~al.(2023)Bubeck, Chandrasekaran, Eldan, Gehrke, Horvitz,
  Kamar, Lee, Lee, Li, Lundberg, et~al.]{bubeck2023sparks}
Bubeck, S., Chandrasekaran, V., Eldan, R., Gehrke, J., Horvitz, E., Kamar, E.,
  Lee, P., Lee, Y.~T., Li, Y., Lundberg, S., et~al.
\newblock Sparks of artificial general intelligence: Early experiments with
  gpt-4.
\newblock \emph{arXiv preprint arXiv:2303.12712}, 2023.

\bibitem[Carroll et~al.(2019)Carroll, Shah, Ho, Griffiths, Seshia, Abbeel, and
  Dragan]{carroll2019utility}
Carroll, M., Shah, R., Ho, M.~K., Griffiths, T., Seshia, S., Abbeel, P., and
  Dragan, A.
\newblock On the utility of learning about humans for human-ai coordination.
\newblock \emph{Advances in neural information processing systems}, 32, 2019.

\bibitem[Chabris et~al.(2010)Chabris, Laibson, and
  Schuldt]{chabris2010intertemporal}
Chabris, C.~F., Laibson, D.~I., and Schuldt, J.~P.
\newblock Intertemporal choice.
\newblock In \emph{Behavioural and experimental economics}, pp.\  168--177.
  Springer, 2010.

\bibitem[Chan et~al.(2021)Chan, Critch, and Dragan]{chan2021human}
Chan, L., Critch, A., and Dragan, A.
\newblock Human irrationality: both bad and good for reward inference.
\newblock \emph{arXiv preprint arXiv:2111.06956}, 2021.

\bibitem[Chiang \& Lee(2023)Chiang and Lee]{chiang2023can}
Chiang, C.-H. and Lee, H.-y.
\newblock Can large language models be an alternative to human evaluations?
\newblock \emph{arXiv preprint arXiv:2305.01937}, 2023.

\bibitem[Coletta et~al.(2022)Coletta, Moulin, Vyetrenko, and
  Balch]{coletta2022learning}
Coletta, A., Moulin, A., Vyetrenko, S., and Balch, T.
\newblock Learning to simulate realistic limit order book markets from data as
  a world agent.
\newblock In \emph{Proceedings of the Third ACM International Conference on AI
  in Finance}, pp.\  428--436, 2022.

\bibitem[Coletta et~al.(2023)Coletta, Vyetrenko, and Balch]{coletta2023k}
Coletta, A., Vyetrenko, S., and Balch, T.
\newblock K-{SHAP}: Policy clustering algorithm for anonymous multi-agent
  state-action pairs.
\newblock In \emph{Proceedings of the 40th International Conference on Machine
  Learning}, pp.\  6343--6363. PMLR, 23--29 Jul 2023.
\newblock URL \url{https://proceedings.mlr.press/v202/coletta23a.html}.

\bibitem[Correia \& Alexandre(2023)Correia and Alexandre]{correia2023survey}
Correia, A. and Alexandre, L.~A.
\newblock A survey of demonstration learning.
\newblock \emph{arXiv preprint arXiv:2303.11191}, 2023.

\bibitem[Day et~al.(2000)Day, Mensink, and O'Sullivan]{day2000patterns}
Day, V., Mensink, D., and O'Sullivan, M.
\newblock Patterns of academic procrastination.
\newblock \emph{Journal of College Reading and Learning}, 30\penalty0
  (2):\penalty0 120--134, 2000.

\bibitem[Desposato(2015)]{desposato2015ethics}
Desposato, S.
\newblock \emph{Ethics and experiments: Problems and solutions for social
  scientists and policy professionals}.
\newblock Routledge, 2015.

\bibitem[Dong et~al.(2023)Dong, Dwarakanath, and Vyetrenko]{dong2023analyzing}
Dong, J., Dwarakanath, K., and Vyetrenko, S.
\newblock Analyzing the impact of tax credits on households in simulated
  economic systems with learning agents.
\newblock \emph{arXiv preprint arXiv:2311.17252}, 2023.

\bibitem[Dragan et~al.(2015)Dragan, Bauman, Forlizzi, and
  Srinivasa]{dragan2015effects}
Dragan, A.~D., Bauman, S., Forlizzi, J., and Srinivasa, S.~S.
\newblock Effects of robot motion on human-robot collaboration.
\newblock In \emph{Proceedings of the Tenth Annual ACM/IEEE International
  Conference on Human-Robot Interaction}, pp.\  51--58, 2015.

\bibitem[Farmer \& Foley(2009)Farmer and Foley]{farmer2009economy}
Farmer, J.~D. and Foley, D.
\newblock The economy needs agent-based modelling.
\newblock \emph{Nature}, 460\penalty0 (7256):\penalty0 685--686, 2009.

\bibitem[Fei et~al.(2022)Fei, Lu, Gao, Yang, Huo, Wen, Lu, Song, Gao, Xiang,
  et~al.]{fei2022towards}
Fei, N., Lu, Z., Gao, Y., Yang, G., Huo, Y., Wen, J., Lu, H., Song, R., Gao,
  X., Xiang, T., et~al.
\newblock Towards artificial general intelligence via a multimodal foundation
  model.
\newblock \emph{Nature Communications}, 13\penalty0 (1):\penalty0 3094, 2022.

\bibitem[Frederick et~al.(2002)Frederick, Loewenstein, and
  O’donoghue]{frederick2002time}
Frederick, S., Loewenstein, G., and O’donoghue, T.
\newblock Time discounting and time preference: A critical review.
\newblock \emph{Journal of economic literature}, 40\penalty0 (2):\penalty0
  351--401, 2002.

\bibitem[Frederiks et~al.(2015)Frederiks, Stenner, and
  Hobman]{frederiks2015household}
Frederiks, E.~R., Stenner, K., and Hobman, E.~V.
\newblock Household energy use: Applying behavioural economics to understand
  consumer decision-making and behaviour.
\newblock \emph{Renewable and Sustainable Energy Reviews}, 41:\penalty0
  1385--1394, 2015.

\bibitem[Gao et~al.(2023)Gao, Lan, Lu, Mao, Piao, Wang, Jin, and Li]{gao2023s}
Gao, C., Lan, X., Lu, Z., Mao, J., Piao, J., Wang, H., Jin, D., and Li, Y.
\newblock S$^3$: Social-network simulation system with large language
  model-empowered agents.
\newblock \emph{arXiv preprint arXiv:2307.14984}, 2023.

\bibitem[Green et~al.(1997)Green, Myerson, and McFadden]{green1997rate}
Green, L., Myerson, J., and McFadden, E.
\newblock Rate of temporal discounting decreases with amount of reward.
\newblock \emph{Memory \& cognition}, 25:\penalty0 715--723, 1997.

\bibitem[Guo et~al.(2023)Guo, Zhang, Wang, Jiang, Nie, Ding, Yue, and
  Wu]{guo2023close}
Guo, B., Zhang, X., Wang, Z., Jiang, M., Nie, J., Ding, Y., Yue, J., and Wu, Y.
\newblock How close is chatgpt to human experts? comparison corpus, evaluation,
  and detection.
\newblock \emph{arXiv preprint arXiv:2301.07597}, 2023.

\bibitem[G{\"u}th et~al.(1982)G{\"u}th, Schmittberger, and
  Schwarze]{guth1982experimental}
G{\"u}th, W., Schmittberger, R., and Schwarze, B.
\newblock An experimental analysis of ultimatum bargaining.
\newblock \emph{Journal of economic behavior \& organization}, 3\penalty0
  (4):\penalty0 367--388, 1982.

\bibitem[Ha \& Schmidhuber(2018)Ha and Schmidhuber]{ha2018world}
Ha, D. and Schmidhuber, J.
\newblock World models.
\newblock \emph{arXiv preprint arXiv:1803.10122}, 2018.

\bibitem[Hagendorff et~al.(2022)Hagendorff, Fabi, and
  Kosinski]{hagendorff2022machine}
Hagendorff, T., Fabi, S., and Kosinski, M.
\newblock Machine intuition: Uncovering human-like intuitive decision-making in
  gpt-3.5.
\newblock \emph{arXiv preprint arXiv:2212.05206}, 2022.

\bibitem[Hendrycks et~al.(2021)Hendrycks, Burns, Kadavath, Arora, Basart, Tang,
  Song, and Steinhardt]{hendrycks2021measuring}
Hendrycks, D., Burns, C., Kadavath, S., Arora, A., Basart, S., Tang, E., Song,
  D., and Steinhardt, J.
\newblock Measuring mathematical problem solving with the math dataset.
\newblock \emph{arXiv preprint arXiv:2103.03874}, 2021.

\bibitem[Hester et~al.(2018)Hester, Vecerik, Pietquin, Lanctot, Schaul, Piot,
  Horgan, Quan, Sendonaris, Osband, et~al.]{hester2018deep}
Hester, T., Vecerik, M., Pietquin, O., Lanctot, M., Schaul, T., Piot, B.,
  Horgan, D., Quan, J., Sendonaris, A., Osband, I., et~al.
\newblock Deep q-learning from demonstrations.
\newblock In \emph{Proceedings of the AAAI conference on artificial
  intelligence}, volume~32, 2018.

\bibitem[Ho \& Ermon(2016)Ho and Ermon]{ho2016generative}
Ho, J. and Ermon, S.
\newblock Generative adversarial imitation learning.
\newblock \emph{Advances in neural information processing systems}, 29, 2016.

\bibitem[Ho et~al.(2022)Ho, Schmid, and Yun]{ho2022large}
Ho, N., Schmid, L., and Yun, S.-Y.
\newblock Large language models are reasoning teachers.
\newblock \emph{arXiv preprint arXiv:2212.10071}, 2022.

\bibitem[Horton(2023)]{horton2023large}
Horton, J.~J.
\newblock Large language models as simulated economic agents: What can we learn
  from homo silicus?
\newblock Technical report, National Bureau of Economic Research, 2023.

\bibitem[Houser \& McCabe(2013)Houser and McCabe]{HOUSER201419}
Houser, D. and McCabe, K.
\newblock \emph{Neuroeconomics: Chapter 2. Experimental Economics and
  Experimental Game Theory}.
\newblock Elsevier Inc. Chapters, 2013.

\bibitem[Hu et~al.(2022)Hu, Corrado, Griffiths, Murez, Gurau, Yeo, Kendall,
  Cipolla, and Shotton]{hu2022model}
Hu, A., Corrado, G., Griffiths, N., Murez, Z., Gurau, C., Yeo, H., Kendall, A.,
  Cipolla, R., and Shotton, J.
\newblock Model-based imitation learning for urban driving.
\newblock \emph{Advances in Neural Information Processing Systems},
  35:\penalty0 20703--20716, 2022.

\bibitem[Huang \& Chang(2022)Huang and Chang]{huang2022towards}
Huang, J. and Chang, K. C.-C.
\newblock Towards reasoning in large language models: A survey.
\newblock \emph{arXiv preprint arXiv:2212.10403}, 2022.

\bibitem[Hussein et~al.(2017)Hussein, Gaber, Elyan, and
  Jayne]{hussein2017imitation}
Hussein, A., Gaber, M.~M., Elyan, E., and Jayne, C.
\newblock Imitation learning: A survey of learning methods.
\newblock \emph{ACM Computing Surveys (CSUR)}, 50\penalty0 (2):\penalty0 1--35,
  2017.

\bibitem[Kahneman \& Tversky(1979)Kahneman and Tversky]{kahneman1979prospect}
Kahneman, D. and Tversky, A.
\newblock Prospect theory: An analysis of decision under risk.
\newblock \emph{Econometrica}, 47\penalty0 (2):\penalty0 263--292, 1979.

\bibitem[Kojima et~al.(2022)Kojima, Gu, Reid, Matsuo, and
  Iwasawa]{kojima2022large}
Kojima, T., Gu, S.~S., Reid, M., Matsuo, Y., and Iwasawa, Y.
\newblock Large language models are zero-shot reasoners.
\newblock \emph{Advances in neural information processing systems},
  35:\penalty0 22199--22213, 2022.

\bibitem[Korinek(2023)]{korinek2023language}
Korinek, A.
\newblock Language models and cognitive automation for economic research.
\newblock Technical report, National Bureau of Economic Research, 2023.

\bibitem[Krawczyk(2018)]{KRAWCZYK2018283}
Krawczyk, D.~C.
\newblock Social cognition.
\newblock In \emph{Reasoning: The neuroscience of how we think}, pp.\
  283--311. Elsevier, 2018.

\bibitem[Kuhn(1964)]{kuhn1964structure}
Kuhn, T.~S.
\newblock \emph{The structure of scientific revolutions}.
\newblock University of Chicago press, 1964.

\bibitem[Kwon et~al.(2022)Kwon, Xie, Bullard, and Sadigh]{kwon2022reward}
Kwon, M., Xie, S.~M., Bullard, K., and Sadigh, D.
\newblock Reward design with language models.
\newblock In \emph{The Eleventh International Conference on Learning
  Representations}, 2022.

\bibitem[Laibson(1994)]{laibson1994hyperbolic}
Laibson, D.~I.
\newblock \emph{Hyperbolic discounting and consumption}.
\newblock PhD thesis, Massachusetts Institute of Technology, 1994.

\bibitem[Langer(1975)]{langer1975illusion}
Langer, E.~J.
\newblock The illusion of control.
\newblock \emph{Journal of personality and social psychology}, 32\penalty0
  (2):\penalty0 311, 1975.

\bibitem[LeBaron(2006)]{lebaron2006agent}
LeBaron, B.
\newblock Agent-based computational finance.
\newblock \emph{Handbook of computational economics}, 2:\penalty0 1187--1233,
  2006.

\bibitem[Liu et~al.(2022)Liu, Dwarakanath, and Vyetrenko]{liu2022biased}
Liu, P., Dwarakanath, K., and Vyetrenko, S.~S.
\newblock Biased or limited: Modeling sub-rational human investors in financial
  markets.
\newblock \emph{arXiv preprint arXiv:2210.08569}, 2022.

\bibitem[Matsuo et~al.(2022)Matsuo, LeCun, Sahani, Precup, Silver, Sugiyama,
  Uchibe, and Morimoto]{matsuo2022deep}
Matsuo, Y., LeCun, Y., Sahani, M., Precup, D., Silver, D., Sugiyama, M.,
  Uchibe, E., and Morimoto, J.
\newblock Deep learning, reinforcement learning, and world models.
\newblock \emph{Neural Networks}, 152:\penalty0 267--275, 2022.

\bibitem[Min et~al.(2021)Min, Ross, Sulem, Veyseh, Nguyen, Sainz, Agirre,
  Heintz, and Roth]{min2021recent}
Min, B., Ross, H., Sulem, E., Veyseh, A. P.~B., Nguyen, T.~H., Sainz, O.,
  Agirre, E., Heintz, I., and Roth, D.
\newblock Recent advances in natural language processing via large pre-trained
  language models: A survey.
\newblock \emph{ACM Computing Surveys}, 2021.

\bibitem[Mischel \& Ebbesen(1970)Mischel and Ebbesen]{mischel1970attention}
Mischel, W. and Ebbesen, E.~B.
\newblock Attention in delay of gratification.
\newblock \emph{Journal of personality and social psychology}, 16\penalty0
  (2):\penalty0 329, 1970.

\bibitem[Ng et~al.(2000)Ng, Russell, et~al.]{ng2000algorithms}
Ng, A.~Y., Russell, S., et~al.
\newblock Algorithms for inverse reinforcement learning.
\newblock In \emph{Icml}, volume~1, pp.\ ~2, 2000.

\bibitem[O'Donoghue \& Rabin(1999)O'Donoghue and Rabin]{o1999doing}
O'Donoghue, T. and Rabin, M.
\newblock Doing it now or later.
\newblock \emph{American economic review}, 89\penalty0 (1):\penalty0 103--124,
  1999.

\bibitem[OpenAI(2023{\natexlab{a}})]{openai2023gpt4}
OpenAI.
\newblock Gpt-4 technical report, 2023{\natexlab{a}}.

\bibitem[OpenAI(2023{\natexlab{b}})]{openai_chatgpt}
OpenAI.
\newblock Introducing chatgpt, 2023{\natexlab{b}}.
\newblock URL \url{https://openai.com/blog/chatgpt}.
\newblock Accessed on: 2023-05-10.

\bibitem[Park et~al.(2022)Park, Popowski, Cai, Morris, Liang, and
  Bernstein]{park2022social}
Park, J.~S., Popowski, L., Cai, C., Morris, M.~R., Liang, P., and Bernstein,
  M.~S.
\newblock Social simulacra: Creating populated prototypes for social computing
  systems.
\newblock In \emph{Proceedings of the 35th Annual ACM Symposium on User
  Interface Software and Technology}, pp.\  1--18, 2022.

\bibitem[Park et~al.(2023)Park, O'Brien, Cai, Morris, Liang, and
  Bernstein]{park2023generative}
Park, J.~S., O'Brien, J.~C., Cai, C.~J., Morris, M.~R., Liang, P., and
  Bernstein, M.~S.
\newblock Generative agents: Interactive simulacra of human behavior.
\newblock \emph{arXiv preprint arXiv:2304.03442}, 2023.

\bibitem[Pollak(1968)]{pollak1968consistent}
Pollak, R.~A.
\newblock Consistent planning.
\newblock \emph{The Review of Economic Studies}, 35\penalty0 (2):\penalty0
  201--208, 1968.

\bibitem[Reddy et~al.(2018)Reddy, Dragan, and Levine]{reddy2018you}
Reddy, S., Dragan, A., and Levine, S.
\newblock Where do you think you're going?: Inferring beliefs about dynamics
  from behavior.
\newblock \emph{Advances in Neural Information Processing Systems}, 31, 2018.

\bibitem[Reynolds \& McDonell(2021)Reynolds and McDonell]{reynolds2021prompt}
Reynolds, L. and McDonell, K.
\newblock Prompt programming for large language models: Beyond the few-shot
  paradigm.
\newblock In \emph{Extended Abstracts of the 2021 CHI Conference on Human
  Factors in Computing Systems}, pp.\  1--7, 2021.

\bibitem[Rothblum et~al.(1986)Rothblum, Solomon, and
  Murakami]{rothblum1986affective}
Rothblum, E.~D., Solomon, L.~J., and Murakami, J.
\newblock Affective, cognitive, and behavioral differences between high and low
  procrastinators.
\newblock \emph{Journal of counseling psychology}, 33\penalty0 (4):\penalty0
  387, 1986.

\bibitem[Sanfey et~al.(2003)Sanfey, Rilling, Aronson, Nystrom, and
  Cohen]{sanfey2003neural}
Sanfey, A.~G., Rilling, J.~K., Aronson, J.~A., Nystrom, L.~E., and Cohen, J.~D.
\newblock The neural basis of economic decision-making in the ultimatum game.
\newblock \emph{Science}, 300\penalty0 (5626):\penalty0 1755--1758, 2003.

\bibitem[Schramowski et~al.(2022)Schramowski, Turan, Andersen, Rothkopf, and
  Kersting]{schramowski2022large}
Schramowski, P., Turan, C., Andersen, N., Rothkopf, C.~A., and Kersting, K.
\newblock Large pre-trained language models contain human-like biases of what
  is right and wrong to do.
\newblock \emph{Nature Machine Intelligence}, 4\penalty0 (3):\penalty0
  258--268, 2022.

\bibitem[Sejnowski(2023)]{sejnowski2023large}
Sejnowski, T.~J.
\newblock Large language models and the reverse turing test.
\newblock \emph{Neural computation}, 35\penalty0 (3):\penalty0 309--342, 2023.

\bibitem[Shah et~al.(2023)Shah, Marupudi, Koenen, Bhardwaj, and
  Varma]{shah2023numeric}
Shah, R.~S., Marupudi, V., Koenen, R., Bhardwaj, K., and Varma, S.
\newblock Numeric magnitude comparison effects in large language models.
\newblock \emph{arXiv preprint arXiv:2305.10782}, 2023.

\bibitem[Sharot et~al.(2007)Sharot, Riccardi, Raio, and
  Phelps]{sharot2007neural}
Sharot, T., Riccardi, A.~M., Raio, C.~M., and Phelps, E.~A.
\newblock Neural mechanisms mediating optimism bias.
\newblock \emph{Nature}, 450\penalty0 (7166):\penalty0 102--105, 2007.

\bibitem[Simon(1955)]{simon1955behavioral}
Simon, H.~A.
\newblock A behavioral model of rational choice.
\newblock \emph{The quarterly journal of economics}, pp.\  99--118, 1955.

\bibitem[Simon(1997)]{simon1997models}
Simon, H.~A.
\newblock \emph{Models of bounded rationality: Empirically grounded economic
  reason}, volume~3.
\newblock MIT press, 1997.

\bibitem[Solomon \& Rothblum(1984)Solomon and Rothblum]{solomon1984academic}
Solomon, L.~J. and Rothblum, E.~D.
\newblock Academic procrastination: Frequency and cognitive-behavioral
  correlates.
\newblock \emph{Journal of counseling psychology}, 31\penalty0 (4):\penalty0
  503, 1984.

\bibitem[Sutton \& Barto(2018)Sutton and Barto]{sutton2018reinforcement}
Sutton, R.~S. and Barto, A.~G.
\newblock \emph{Reinforcement learning: An introduction}.
\newblock MIT press, 2018.

\bibitem[Thaler(1988)]{thaler1988anomalies}
Thaler, R.~H.
\newblock Anomalies: The ultimatum game.
\newblock \emph{Journal of economic perspectives}, 2\penalty0 (4):\penalty0
  195--206, 1988.

\bibitem[Thaler et~al.(1997)Thaler, Tversky, Kahneman, and
  Schwartz]{thaler1997effect}
Thaler, R.~H., Tversky, A., Kahneman, D., and Schwartz, A.
\newblock The effect of myopia and loss aversion on risk taking: An
  experimental test.
\newblock \emph{The quarterly journal of economics}, 112\penalty0 (2):\penalty0
  647--661, 1997.

\bibitem[Tversky \& Kahneman(1992)Tversky and Kahneman]{tversky1992advances}
Tversky, A. and Kahneman, D.
\newblock Advances in prospect theory: Cumulative representation of
  uncertainty.
\newblock \emph{Journal of Risk and uncertainty}, 5:\penalty0 297--323, 1992.

\bibitem[Wei et~al.(2022{\natexlab{a}})Wei, Tay, Bommasani, Raffel, Zoph,
  Borgeaud, Yogatama, Bosma, Zhou, Metzler, et~al.]{wei2022emergent}
Wei, J., Tay, Y., Bommasani, R., Raffel, C., Zoph, B., Borgeaud, S., Yogatama,
  D., Bosma, M., Zhou, D., Metzler, D., et~al.
\newblock Emergent abilities of large language models.
\newblock \emph{Transactions on Machine Learning Research}, 2022{\natexlab{a}}.

\bibitem[Wei et~al.(2022{\natexlab{b}})Wei, Wang, Schuurmans, Bosma, Xia, Chi,
  Le, Zhou, et~al.]{wei2022chain}
Wei, J., Wang, X., Schuurmans, D., Bosma, M., Xia, F., Chi, E., Le, Q.~V.,
  Zhou, D., et~al.
\newblock Chain-of-thought prompting elicits reasoning in large language
  models.
\newblock \emph{Advances in Neural Information Processing Systems},
  35:\penalty0 24824--24837, 2022{\natexlab{b}}.

\bibitem[Williams et~al.(2023)Williams, Hosseinichimeh, Majumdar, and
  Ghaffarzadegan]{williams2023epidemic}
Williams, R., Hosseinichimeh, N., Majumdar, A., and Ghaffarzadegan, N.
\newblock Epidemic modeling with generative agents.
\newblock \emph{arXiv preprint arXiv:2307.04986}, 2023.

\bibitem[Ziems et~al.(2023)Ziems, Held, Shaikh, Chen, Zhang, and
  Yang]{ziems2023can}
Ziems, C., Held, W., Shaikh, O., Chen, J., Zhang, Z., and Yang, D.
\newblock Can large language models transform computational social science?
\newblock \emph{arXiv preprint arXiv:2305.03514}, 2023.

\end{thebibliography}
\bibliographystyle{icml2024}

\clearpage
\appendix
\section{Experimental Details}
We report additional details about the experimental results in order to improve the reproducibility and usage of our framework.
In particular, for each experimental game we report the prompt and the answers from the LLMs, including some analysis on the distribution of results. 

\subsection{Implementation}
We implement our framework using Python3.9. The DQN algorithm is implemented using PyTorch and is composed of 3 full connected layers, where the input is a one-hot encoding of the state and the output represents the probability distribution over the actions space $\mathcal{A}$. We use OpenAI GPT-4~\cite{openai2023gpt4} API with model id. \textit{"gpt-4-0613"}. Where not otherwise stated, we generate 10  synthetic observations for each state, using the OpenAI python API for Chat Completions based on GPT-4 with temperature equal to 0.5 and a maximum of 5 output tokens. We defined the messages using the \textbf{system role}, to set the behavior of the LLMs, modifying its personality and providing specific instructions about how it should behave, and the \textbf{user role} to specify the environment state in which the LLM should take an action. Notice that, we specifically generate few demonstrations as the considered games are relatively simple (i.e, small state-action space). The few demonstrations are also used to demonstrate the ability of our framework to learn sub-rational policies from few data samples. 

\subsection{Ultimatum Game}

\paragraph{\textbf{Human behavior.}}
In our first test we evaluate how the model align with human behavior in the ultimatum game of Section~\ref{sec:ultimatum}, possibly rejecting too low offers. In such a case we define the API message as follow in Table~\ref{tab:human_beh_ultimatum}:

\begin{table}[h]
\centering
\begin{tabular}{|c|p{0.8\linewidth}|}
\hline
\textbf{Role} & \textbf{Content} \\
\hline
System & Impersonate \textbf{a human} called Jerry \\ \hline
User & {\em John receives \$T, and he proposes to offer Jerry x and keep T-x  for himself. If Jerry accepts they both get the agreed amount, but if Jerry rejects they both receive nothing. Continue the sentence: 'Jerry decides to'} \\
\hline
\end{tabular}
    \caption{Ultimatum Game: Human Behavior prompt.}
    \label{tab:human_beh_ultimatum}
\end{table}

Where we set the amount of money $T=10$ and we vary the state between the 10 different possible splits $x \in \{0, \ldots, T\}$. We then convert each prompt and answer into a state-action (s, a) pair. We generate 10 synthetic observations for each state $x$.
In Table~\ref{tab:human_prob_ultimatum} we report the rate at which the "human" LLM accept the proposed amount. All the raw answers are reported in table~\ref{tab:human_res}.

\begin{table}[t]
\centering
\resizebox{0.5\textwidth}{!}{
    \begin{tabular}{|l|lllllllllll|}
\toprule
\textbf{Proposed Amount }&  \$0  &   \$1  &   \$2  &   \$3  &   \$4  &   \$5  &   \$6  &   \$7  &   \$8 &   \$9  &   \$10 \\
\midrule
\textbf{Acceptance Rate} &  0\% &  0\% &  20\% &  90\% &  100\% &  100\% & 100\% &  100\% &  100\% &  100\% &  100\% \\
\bottomrule
\end{tabular}
}
    \caption{Ultimatum Game: Human Behavior acceptance rate.}
    \label{tab:human_prob_ultimatum}
\end{table}

\newpage
\paragraph{\textbf{Fair behavior.}}
We evaluate how the model align w.r.t. a more fair human behavior, in the ultimatum game of Section~\ref{sec:ultimatum}, possibly rejecting unfair splits. In such a case we define the API message as follow in Table~\ref{tab:human_fair_ultimatum}:

\begin{table}[h]
\centering
\begin{tabular}{|c|p{0.8\linewidth}|}
\hline
\textbf{Role} & \textbf{Content} \\
\hline
System & Impersonate Jerry \textbf{who cares a lot about fairness} \\ \hline
User & {\em John receives \$T, and he proposes to offer Jerry x and keep T-x  for himself. If Jerry accepts they both get the agreed amount, but if Jerry rejects they both receive nothing. Continue the sentence: 'Jerry decides to'} \\
\hline
\end{tabular}
    \caption{Ultimatum Game: Fair Behavior prompt.}
    \label{tab:human_fair_ultimatum}
\end{table}

Also in this case we set the amount of money $T=10$ and we vary the state between the 10 different possible splits $x \in \{0, \ldots, T\}$. We then convert each prompt and answer into a state-action (s, a) pair. We generate 10 synthetic observations for each state $x$.
In Table~\ref{tab:human_fair__prob_ultimatum} we report the rate at which the "fair" LLM accept the proposed amount. All the raw answers are reported in table~\ref{tab:human_fair}.

\begin{table}[h]
\centering
\resizebox{0.5\textwidth}{!}{
    \begin{tabular}{|l|lllllllllll|}
\toprule
\textbf{Proposed Amount }&  \$0  &   \$1  &   \$2  &   \$3  &   \$4  &   \$5  &   \$6  &   \$7  &   \$8 &   \$9  &   \$10 \\
\midrule
\textbf{Acceptance Rate} &  0\% &  0\% &  0\% &  0\% &  0\% &  40\% &  0\% &  0\% &  0\% &  0\% &  100\% \\
\bottomrule
\end{tabular}
}
    \caption{Ultimatum Game: Fair Behavior acceptance rate.}
    \label{tab:human_fair__prob_ultimatum}
\end{table}

Surprisingly, even if the LLMs clearly see a fair split as one around 50:50, the LLMs is also accepting a completely unfair split in its favor (i.e., Jerry takes all the money). 

\paragraph{\textbf{Discussion.}}
We note that in our experiment we attempt to use a common name like Jerry to avoid possible biases: the work in \cite{aher2023using} shows that LLMs can have different results when conditioned on names from different ethnicity or gender. While biases could be bad in machine learning models, in our cases this help to simulate and model different human behaviors.  We will show in the next experiment as a age biases could be extremely useful to model a famous physiological game, which is otherwise unclear how to model using classical RL approaches. 

\subsection{Stanford Marshmallow Experiment}
For the Stanford marshmallow experiment we use the prompt define in Section~\ref{sec:stanford_exp}. We here report more extensive results, simulating child with age $Y \in \{2, 3, 4, 5\}$, and using as waiting times both \textit{15 minutes} and \textit{2 hours}. We evaluate how the model align with child behavior, possibly accepting immediate rewards. In such a case we define the API message as follow in Table~\ref{tab:stanford_prompt}:

\begin{table}[h]
\centering
\resizebox{0.5\textwidth}{!}{
\begin{tabular}{|c|p{0.8\linewidth}|}
\hline
\textbf{Role} & \textbf{Content} \\
\hline
System & You are Janice a Y years old child. \\ \hline
User (2 hours) & {\em Janice is offered to get one candy now, or to wait for \textbf{2 more hours} to eventually get two candies. Continue the following sentence 'Janice decides to'} \\
\hline
User (15 min) & {\em Janice is offered to get one candy now, or to wait for \textbf{15 minutes} to eventually get two candies. Continue the following sentence 'Janice decides to'} \\
\hline
\end{tabular}
}
    \caption{Stanford marshmallow experiment prompt.}
    \label{tab:stanford_prompt}
\end{table}

We generate 10 synthetic observations for each age $Y$, limiting the model to generate 5 tokens.  In Table~\ref{tab:stanford_prob} we report the probability of waiting, computed as the percentage of observations where the LLM impersonate a waiting child. When we use "15 minutes" waiting time as in the original experiment, we notice that the model is more likely to wait, which may indicate a different human preferences w.r.t. original paper dated more than 50 years ago. Additionally, although it seems plausible that the percentage of children waiting increase as the waiting time decreases, we question the extent to which the model is overly rational. A two-year-old child may not possess enough awareness of the concept of time to distinguish and express a preference for 15 minutes over 2 hours.
\\
We report all the raw LLM answers in Table~\ref{tab:stanford_answers} and Table~\ref{tab:stanford_answers_15min} for 2 hours and 15minutes waiting times, respectively. 

\begin{table}[h]
    \centering    
    \resizebox{0.5\textwidth}{!}{
\begin{tabular}{c|c|c}
         \hline
         Age of child & Waiting prob. (2 hr.) & Waiting prob. (15 min)  \\ \hline
         2 years  & $0.0$ & $0.2$ \\
         3 years & $0.2$ & $0.8$ \\
         4 years & $1.0$ & $1.0$ \\
         5 years & $1.0$ & $1.0$ \\ \hline
    \end{tabular}
    }
    \caption{Stanford marshmallow experiment: Waiting probability in LLM demonstrations. }
    \label{tab:stanford_prob}
\end{table}

\subsection{Double or Nothing Gamble}
Following \cite{tversky1992advances}, we implement the prospect theory model with the following biased value function and probability weighting function:

\noindent\textbf{Biased Value Function.} Empirical studies of human decisions often demonstrate loss aversion, showing that humans feel losses more than two times greater than the equivalent gains. To capture this behavior,~\cite{tversky1992advances} introduce the value function as
\begin{equation}
    v(x) =
    \begin{cases}
      x^\alpha & \text{if $x \geq 0$}\\
      -\lambda (-x)^\alpha & \text{if $x < 0$}
    \end{cases}.
    \label{eq:prospect_utility}
\end{equation}
where the parameters are estimated from real human data ($\alpha=0.88$ and $\lambda=2.25$).
The value function is concave for gains, convex for losses, and steeper for losses than gains. Thus, humans tend to be \textit{risk-averse} over gains, preferring \$450 with probability $1$, over \$1000 with probability 0.5; and \textit{risk-seeking} over losses, preferring to lose -\$1000 with probability 0.5 rather than having a certain loss of -\$450.

\noindent\textbf{Probability Weighting Function.}
\noindent Another aspect of the prospect theory is that humans do not perceive probability in a linear fashion. When addressing any uncertainty, humans consider the natural boundaries of probability, impossibility and certainty, as the two reference points. The impact of a change in probability diminishes with its distance to the reference points. Considering a 0.1 increase in the probability of winning a prize, a change from 0 to 0.1 chance to win has more impact than a change from 0.45 to 0.55. To model this distortion in probability seen in humans,~ \cite{tversky1992advances} introduce a non-linear weighting function as
\begin{equation}\label{eq:prospect_prob}
    w(p) = \frac{p^\delta}{(p^\delta + (1 - p)^\delta)^{1/\delta}}
\end{equation}
where $p$ is the probability of a potential outcome, and $\delta$ is estimated to be 0.61, 0.69 for gains, losses respectively. 

In addition to the prompt in~\cref{sec:prospect_exp}, we collect the LLM demonstration from GPT-4 using the API message in ~\cref{tab:prospect_prompt}.

\begin{table}[h]
\centering
\resizebox{0.5\textwidth}{!}{
\begin{tabular}{|c|p{0.8\linewidth}|}
\hline
\textbf{Role} & \textbf{Content} \\
\hline
System & You impersonate an average human. \\ \hline
User (winner) & {\em You won an initial \$5 bet against Tom. Before you collect your reward, you can choose to add a second bet of \$5 with {$0.5+\epsilon$} probability to win. If you win the second bet you will double your gain from the initial bet, but if you loose you will gain nothing. Continue the following sentence `You decides to accept/reject the second bet'} \\
\hline
User (loser) & {\em You lost an initial \$5 bet against Tom. Before you pay your debt, you can choose to add a second bet of \$5 with {$0.5-\epsilon$} probability to win. If you win the second bet you will recover your loss from the initial bet, but if you loose you will double your loss. Continue the following sentence `You decides to accept/reject the second bet'} \\
\hline
\end{tabular}}
    \caption{Double or nothing gamble prompt.}
    \label{tab:prospect_prompt}
\end{table}

\noindent\textbf{Discussion.}
While in theory fully rational human should always aim to maximize their expected rewards, we observe that GPT-4 prompted as "rational human" sometimes choose to play safe and deny the second bet as winner. In fact, over the past decades the human society has developed more concerns on risk, and a low-risk strategy will be recognized as good decision even if the expected utility is mediocre. This interesting results indicates that the LLMs are able to incorporate the changes in perspectives of the society. 

\subsection{Academic Procrastination with Deadlines}
In this experiment, we investigate if GPT-4 exhibits procrastinating behavior observed in students especially in self-paced courses when they have a deliverable due within the next $H$ days.
We try to capture student attributes through their Grade Point Average (GPA), and report the API prompt for this experiment in~\cref{tab:procrastination}. 
\begin{table}[h]
\centering
\begin{tabular}{|c|p{0.8\linewidth}|}
\hline
\textbf{Role} & \textbf{Content} \\
\hline
System & You impersonate a student with GPA $x$, who loves watching movies. GPA measures the students commitment towards their academics. \\ \hline
User & {\em The theatre has a line up of increasingly better movies as days pass. The student has a deadline to submit a course report within the next $y$ days. The student needs to pick one day to write the report, which means they will miss the movie on that day. Continue the sentence `The student writes the report on day '} \\
\hline
\end{tabular}
    \caption{Procrastination experiment prompt.}
    \label{tab:procrastination}
\end{table}

We collect 10 demonstrations for each GPA value $x\in\lbrace1,3,4.5\rbrace$ and each academic deadline $y\in\lbrace4,10\rbrace$. The results and analysis for deadline $H=4$ are reported in~\cref{subsec:procrastination} where we observed that the probability of writing the report on the first day increases as the GPA of the student increases. 
As the deadline is made longer from $4$ days to $10$ days, we observe that the probability of writing the report on the last day decreases as the GPA of the student increases. The trend for writing the report on the first day with GPA does not hold for the longer deadline. We posit that this is potentially due to the inability of LLMs to reason over larger horizons giving a cluster of observations in the intermediate days with few for the extreme days (with 4.5 GPA potentially being less observed in their training data).

\begin{figure}[t]
\centering
\includegraphics[width=0.8\linewidth]{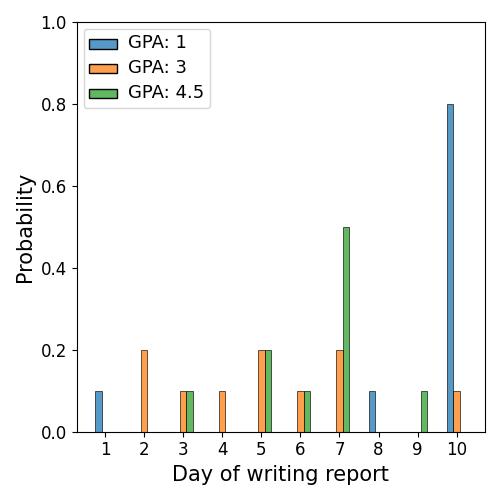}
\caption{
Procrastination experiment: LLM demonstrations when deadline is ten days. 
}\label{fig:procrastination_10}
\end{figure}

The MDP is defined as follows for the RL policy. The state at time $t$ is $s_t=\begin{bmatrix}
    t&f_t
\end{bmatrix}$ where $t\in\lbrace1,2,\cdots,H+1\rbrace$ and $f_t\in\lbrace0,1\rbrace$ is a flag that tracks whether the report was written by time step $t$. The binary action $a_t\in\lbrace0,1\rbrace$ indicates writing ($a_t=1$) or procrastinating ($a_t=0$) on the report at $t$. The reward is given by 
\begin{align*}
    r(s_t,a_t,s_{t+1}) = &-c_ta_t\cdot\mathbbm{1}_{\lbrace f_t=0\rbrace}+\\ & R\cdot\mathbbm{1}_{\lbrace t=H,f_{H+1}=1\rbrace}-\\ &KR\cdot\mathbbm{1}_{\lbrace t=H,f_{H+1}=0\rbrace}
\end{align*}
for $K>1$. The first term says that the student pays a cost $c_t$ at the time $t$ when the report is written the first time. The second term is the final reward given to the student when the report is written before the deadline. The third term is to penalize cases when the student does not write the report within the deadline. When $H=4$, the costs and reward are reported in~\cref{subsec:procrastination} similar to the example in~\cite{o1999doing}. When $H=10$, we use $c_t=t+c_{t-1}\forall t\geq2$ where $c_1=1$ and $R=2c_{10}$.

\noindent\textbf{Discussion.}
Note that we use neutral framing of student attributes other than GPA in our prompt, where we avoid using a name and use neutral pronouns to prevent any biases related to gender, ethnicity, age etc.
Exponential discounting with $\gamma=1$ says it's optimal to write the report on day 1 since they pick $t$ that maximizes $R-c_t$. 
Thus, the student gets to watch all the good movies on day 2 onward. On the other hand, one can show that with quasi-hyperbolic discounting with $\delta=1$ and $\beta=0.4$, the optimal strategy (for sophisticated humans) is to write the report on day 2. This is because the value function for writing the report on day $t$ is $\beta R-c_t$. Whereas, the value function for not writing the report on day $t$ is $\beta(R-c_\tau)$ for $\tau>t$ where $\tau$ is the earliest day of writing the report after day $t$. The student chooses to write a report on day $t$ only if $\beta R-c_t>\beta(R-c_\tau)$ for $\tau$ described earlier. This discrepancy stems from the higher preference for immediate rewards (also called present-bias \cite{benhabib2010present}).

\newpage
\section{Testing for Reasoning in LLM Demonstrations}
We conduct additional experiments to examine whether the LLM demonstrations come from its reasoning capability or the memorization of historical experiment results. In particular, we modify the prompt in~\cref{tab:stanford_prompt} to see if GPT-4 can extrapolate to new scenarios rather than simply mimicking the results of Stanford marshmallow experiment.

\Cref{tab:marshmallow_additional} shows the modified prompts of the Stanford marshmallow experiments along with an example of answers from GPT-4. In the default experiments a 5-year old child is willing to wait for more candies. However, we observe that LLM will not wait as a 5-year old if the delayed reward is replaced with bitter melons or the wait time increases to two years. This may indicate that LLM demonstrations are generated through a reasoning process, which are not simply replicated from the known experiment results in the existing literature.

Meanwhile, we notice that LLMs fail to give reasonable demonstrations for certain scenarios highlighted in Table~\ref{tab:llm_failure}. For example, GPT-4 is willing to wait two hours for an additional bitter melon due to its hallucinations. In addition, GPT-4 sometimes demonstrates poor reasoning on the length of the wait time, and is willing to wait two months for an extra candy. This is possibly related to the well-known issues of LLMs in understanding numerical values~\cite{shah2023numeric}. 

\begin{table}[h]
\centering
\resizebox{0.5\textwidth}{!}{
\begin{tabular}{|c|p{0.8\linewidth}|}
\hline
System & You are Janice a 5 years old child. \\ \hline
User & {\em Janice is offered to get one candy now, or to wait for 2 more hours to eventually get two \textbf{bitter melons}, Continue the following sentence ’Janice decides
to’} \\
Answer & {\em Janice decides to \textbf{take the one candy now}. While two bitter melons might be more, she doesn't really like bitter melons and she wants something sweet right now. She happily takes the candy and enjoys it immediately.} \\
\hline
User & {\em Janice is offered to get one candy now, or to wait for \textbf{2 more years} to eventually get two candies, Continue the following sentence ’Janice decides
to’} \\
Answer & {\em Janice decides to \textbf{take the one candy now} because waiting for two years seemed like a very long time for a five-year-old.} \\
\hline
\end{tabular}
}
    \caption{Modified Stanford marshmallow experiment demonstrating reasoning abilities of LLMs.}
    \label{tab:marshmallow_additional}
\hfill
\centering
\resizebox{0.5\textwidth}{!}{
\begin{tabular}{|c|p{0.8\linewidth}|}
\hline
\textbf{Role} & \textbf{Content} \\\hline
System & You are Janice a  5 years old child \textbf{who doesn't like bitter taste}.\\ 
User & {\em Janice is offered to get one \textbf{bitter melon} now, or to wait for 2 more hours to eventually get two \textbf{bitter melons}, Continue the following sentence ’Janice decides
to’} \\
Answer & {\em Janice decides to \textbf{wait} for two more hours anyway, \textbf{hoping that} someone would want to trade her two bitter melons for something sweet.} \\
\hline
System & You are Janice a 5 years old child. \\ 
User & {\em Janice is offered to get one candy now, or to wait for \textbf{2 more months} to eventually get two candies, Continue the following sentence ’Janice decides
to’} \\
Answer & {\em Janice decides to patiently \textbf{wait} for two more months to eventually get two candies because she loves candies and getting two candies is better than one.} \\
\hline
\end{tabular}
}
\caption{Modified Stanford marshmallow experiment demonstrating hallucinations and inaccurate sensitivity to time.}
    \label{tab:llm_failure}
\end{table}

\begin{table*}[h]
\centering
\resizebox{1\textwidth}{!}{
    \begin{tabular}{|l|llllllllll|l|}
\toprule
$x$ value & 1$^{st}$ Answer & 2$^{nd}$ Answer & 3$^{rd}$ Answer & 4$^{th}$ Answer & 5$^{th}$ Answer& 6$^{th}$ Answer & 7$^{th}$ Answer & 8$^{th}$ Answer & 10$^{th}$ Answer & 10$^{th}$ Answer & \textbf{Acceptance Rate} \\
\midrule
0  &  reject the proposal &        reject the offer &     reject the proposal &  reject the proposal &     reject the offer &       reject the proposal &        reject the offer &  reject the proposal &  reject the offer &          reject the proposal & 0.0 \\
1  &     reject the offer &        reject the offer &        reject the offer &     reject the offer &  reject the proposal &       reject the proposal &     reject the proposal &  reject the proposal &  reject the offer &       reject John's proposal & 0.0 \\
2  &     accept the offer &        reject the offer &        reject the offer &  reject the proposal &     accept the offer &       reject the proposal &        reject the offer &     reject the offer &  reject the offer &          reject the proposal & 0.2 \\
3  &     accept the offer &     accept John's offer &  accept John's proposal &     accept the offer &     reject the offer &       accept the proposal &  accept John's proposal &     accept the offer &  accept the offer &  accept the offer because he & 0.9 \\
4  &     accept the offer &        reject the offer &        accept the offer &     accept the offer &     accept the offer &          accept the offer &        accept the offer &     accept the offer &  accept the offer &             accept the offer & 1.0 \\
5  &  accept the proposal &  accept John's proposal &        accept the offer &     accept the offer &     accept the offer &          accept the offer &  accept John's proposal &     accept the offer &  accept the offer &             accept the offer & 1.0 \\
6  &     accept the offer &        accept the offer &        accept the offer &     accept the offer &     accept the offer &          accept the offer &  accept John's proposal &     accept the offer &  accept the offer &             accept the offer & 1.0 \\
7  &     accept the offer &        accept the offer &     accept the proposal &     accept the offer &     accept the offer &  'Jerry decides to accept &        accept the offer &     accept the offer &  accept the offer &             accept the offer & 1.0 \\
8  &     accept the offer &        accept the offer &        accept the offer &     accept the offer &     accept the offer &          accept the offer &        accept the offer &     accept the offer &  accept the offer &       accept John's proposal & 1.0 \\
9  &     accept the offer &        accept the offer &        accept the offer &     accept the offer &  accept the proposal &          accept the offer &        accept the offer &     accept the offer &  accept the offer &             accept the offer & 1.0 \\
10 &     accept the offer &        accept the offer &        accept the offer &     accept the offer &     accept the offer &  'Jerry decides to accept &        accept the offer &     accept the offer &  accept the offer &             accept the offer & 1.0 \\
\bottomrule
\end{tabular}
}
    \caption{Ultimatum Game: LLM demonstrations for Human Behavior.}
    \label{tab:human_res}
\end{table*}

\begin{table*}[h]
\centering
\resizebox{1\textwidth}{!}{
    \begin{tabular}{|l|llllllllll|l|}
\toprule
$x$ value & 1$^{st}$ Answer & 2$^{nd}$ Answer & 3$^{rd}$ Answer & 4$^{th}$ Answer & 5$^{th}$ Answer& 6$^{th}$ Answer & 7$^{th}$ Answer & 8$^{th}$ Answer & 10$^{th}$ Answer & 10$^{th}$ Answer & \textbf{Acceptance Rate} \\
\midrule
0  &           'reject the proposal. &        reject the proposal &      reject the proposal &             reject the proposal &    reject the proposal. He &         reject the offer &      reject the proposal. " &  reject the proposal. He &             reject the proposal &       reject the proposal. He & 0.0 \\
1  &             reject the proposal &        reject the proposal &         reject the offer &         reject the proposal. He &       reject the offer. He &      reject the offer. " &     reject John's proposal. &  reject the proposal. He &                reject the offer &       reject John's proposal. & 0.0 \\
2  &             reject the proposal &        reject the proposal &  reject the proposal. He &            reject the offer. He &        reject the proposal &         reject the offer &     reject the proposal. He &         reject the offer &         reject the proposal. He &           reject the proposal & 0.0 \\
3  &         reject the proposal. He &       reject the offer. He &         reject the offer &                reject the offer &        reject the proposal &     reject the offer. He &            reject the offer &      reject the proposal &                reject the offer &           reject the offer. " & 0.0 \\
4  &         reject the proposal. He &        reject the proposal &         reject the offer &         reject the proposal. He &     reject the proposal. " &     reject the offer. As &            reject the offer &      reject the proposal &          reject John's proposal &              reject the offer & 0.0 \\
5  &             accept the proposal &        accept the proposal &      reject the proposal &             accept the proposal &        reject the proposal &   reject the proposal. " &         accept the proposal &   reject the offer. Even &                reject the offer &           reject the proposal  & 0.4 \\
6  &                reject the offer &    reject the proposal. He &      reject the proposal &         reject the proposal. He &           reject the offer &     reject the offer. He &  reject the proposition. In &   reject the proposal. " &                reject the offer &              reject the offer & 0.0 \\
7  &             reject the proposal &    reject the proposal. He &      reject the proposal &             reject the proposal &           reject the offer &     reject the offer. He &      reject the proposal. " &         reject the offer &                reject the offer &        reject the proposal. " & 0.0 \\
8  &                reject the offer &    reject the proposal. He &  reject John's proposal. &             reject the proposal &        reject the proposal &  reject John's proposal. &  reject the proposal. Being &     reject the offer. He &         reject John's proposal. &       reject the proposal. He & 0.0 \\
9  &      reject the offer. Although &  reject the offer. Despite &     reject the offer. He &         reject the proposal. He &  reject the offer. Despite &   reject the proposal. " &   reject the offer. Despite &     reject John's offer. &             reject the proposal &  reject the proposal. Despite & 0.0 \\
10 &  accept John's proposal because &        accept the proposal &   accept John's proposal &  accept John's proposal because &        accept John's offer &   accept John's proposal &         accept the proposal &   accept John's proposal &  accept the proposal because it &           accept the proposal & 1.0 \\
\bottomrule
\end{tabular}
}
    \caption{Ultimatum Game: LLM demonstrations for Fair Behavior.}
    \label{tab:human_fair}
\end{table*}

\begin{table}[t]
\centering
\begin{tabular}{|p{0.8\linewidth}|}
\toprule
 \textbf{\textit{2 Hours Waiting}} \\
\midrule
 \textbf{2 Year Old} \\
 \midrule
take the one candy now \\
 ...grab the one candy \\
...grab the one candy \\
take the one candy now \\
take the one candy now \\
take the one candy now \\
take the one candy now \\
take the one candy now \\
take the one candy now \\
take the candy now, \\
\midrule
\textbf{ 3 Year Old} \\
\midrule
...take the one candy \\
...take the one candy \\
wait for 2 more \\
...take one candy now \\
wait for 2 more \\
...take the one candy \\
...take the one candy \\
...take the one candy \\
...take one candy now \\
...take one candy now\\
\midrule
\textbf{ 4 Year Old} \\
\midrule
wait for 2 more \\
wait for 2 more \\
wait for 2 more \\
wait for 2 more \\
wait for 2 more \\
wait for two more hours \\
wait for 2 more \\
wait for 2 more \\
wait for 2 more \\
wait for 2 more\\
\midrule
 \textbf{5 Year Old} \\
\midrule
wait for 2 more \\
wait for 2 more \\
wait for 2 more \\
wait for 2 more \\
wait for 2 more \\
wait for 2 more \\
wait for 2 more \\
wait for 2 more \\
wait for 2 more \\
wait for 2 more\\ 
 \bottomrule
\end{tabular}
    \caption{Stanford Marshmallow Experiment: LLM demonstrations (2 hours).}
    \label{tab:stanford_answers}
\end{table}

\begin{table}[t]
\centering
\begin{tabular}{|p{0.8\linewidth}|}
\toprule
 \textbf{\textit{15 Minutes Waiting}} \\
\midrule
 \textbf{2 Year Old} \\
 \midrule
...grab the one candy \\
take the one candy now \\
...grab the one candy \\
...grab the one candy \\
...grab the candy now \\
take one candy now, \\
take the one candy now \\
take the one candy now \\
grab the one candy now \\
wait for 15 minutes \\
\midrule
\textbf{ 3 Year Old} \\
\midrule
...take one candy now \\
wait for 15 minutes \\
...wait for 15 \\
wait for 15 minutes \\
wait for 15 minutes \\
wait for 15 minutes \\
wait for 15 minutes \\
...grab the candy now \\
wait for 15 minutes \\
wait for 15 minutes\\
\midrule
\textbf{ 4 Year Old} \\
\midrule
wait for 15 minutes \\
wait for 15 minutes \\
wait for 15 minutes \\
wait for 15 minutes \\
...wait for 15 \\
wait for 15 minutes \\
wait for 15 minutes \\
wait for 15 minutes \\
wait for 15 minutes \\
wait for 15 minutes\\
\midrule
 \textbf{5 Year Old} \\
\midrule
wait for 15 minutes \\
wait for 15 minutes \\
wait for 15 minutes \\
wait for 15 minutes \\
wait for 15 minutes \\
wait for 15 minutes \\
wait for 15 minutes \\
wait for 15 minutes \\
wait for 15 minutes \\
wait for 15 minutes \\
 \bottomrule
\end{tabular}
    \caption{Stanford Marshmallow Experiment: LLM demonstrations (15 minutes).}
    \label{tab:stanford_answers_15min}
\end{table}

\end{document}